\documentclass{article} 
\usepackage{iclr2025_conference,times}

\usepackage{amsmath,amsfonts,bm}

\def\secref#1{section~\ref{#1}}

\def\eqref#1{equation~\ref{#1}}

\def\1{\bm{1}}

\DeclareMathAlphabet{\mathsfit}{\encodingdefault}{\sfdefault}{m}{sl}
\SetMathAlphabet{\mathsfit}{bold}{\encodingdefault}{\sfdefault}{bx}{n}

\usepackage[hidelinks]{hyperref}
\usepackage{url}

\newcommand{\equref}[1]{Equation (\ref{#1})}

\title{A Unified View of Delta Parameter Editing in Post-Trained Large-Scale Models}

\iclrfinalcopy
\author{
  Qiaoyu Tang${}^{1,2}$,
  Le Yu${}^{3}$,
  Bowen Yu${}^{3}$~\thanks{~ Corresponding authors.},
  Hongyu Lin${}^{1}$\footnotemark[1],\\
  {\;\bf Keming Lu${}^{3}$,}
  {\bf Yaojie Lu${}^{1}$,}
  {\bf Xianpei Han${}^{1}$,}
  {\bf Le Sun${}^{1}$}
  \\
  ${}^{1}$Chinese Information Processing Laboratory, Institute of Software, Chinese Academy of Sciences\\
  ${}^{2}$University of Chinese Academy of Sciences \\
  ${}^{3}$Alibaba Group \\
  {\tt \{tangqiaoyu2020,hongyu,luyaojie,xianpei,sunle\}@iscas.ac.cn} \\
  {\tt \{chuanyi.yl,yubowen.ybw,lukeming.lkm\}@alibaba-inc.com} \\
}

\usepackage{graphicx}    
\usepackage{caption}     
\usepackage{subcaption}  
\usepackage{wrapfig}
\usepackage{booktabs}
\begin{document}

\maketitle

\begin{abstract}
Post-training has emerged as a crucial paradigm for adapting large-scale pre-trained models to various tasks, whose effects are fully reflected by delta parameters (i.e., the disparity between post-trained and pre-trained parameters). While numerous studies have explored delta parameter properties via operations like pruning, quantization, low-rank approximation, and extrapolation, a unified framework for systematically examining these characteristics has been lacking. In this paper, we propose a novel perspective based on Riemann sum approximation of the loss function to elucidate delta parameter editing operations. Our analysis categorizes existing methods into three classes based on their post-editing performance: competitive, decreased, and improved, explaining how they are expressed by the Riemann sum approximation term and how they alter the model performance. Extensive experiments on both visual and language models, including ViT, LLaMA 3, Qwen 2, and Mistral, corroborate our theoretical findings. Furthermore, we introduce extensions to existing techniques like DARE and BitDelta, highlighting their limitations in leveraging the properties of delta parameters and reorganizing them into general expressions to enhance the applicability and effectiveness of delta parameter editing in post-trained models.
\end{abstract}

\section{Introduction}
\label{section-1}

With the remarkable success of large-scale pre-trained models, post-training has emerged as the de facto standard paradigm for effective adaptations to various tasks \citep{DBLP:journals/corr/abs-2403-14608,DBLP:journals/corr/abs-2402-02242,DBLP:journals/corr/abs-2002-06305,DBLP:journals/corr/abs-2303-18223}. Conceptually, post-training optimizes the parameters of pre-trained backbone on task-specific data, endowing models with diverse abilities like visual recognition \citep{DBLP:conf/nips/ChenGTWSWL22,DBLP:conf/cvpr/0002ZV022}, instruction following \citep{DBLP:conf/nips/RafailovSMMEF23,DBLP:journals/corr/abs-2402-01306}, and mathematical reasoning \citep{DBLP:journals/corr/abs-2308-09583,DBLP:journals/corr/abs-2407-13690}. It has been noted that the impact of post-training is fully manifested in the \textit{delta parameters}, which are defined as the difference between parameters of pre-trained and post-trained models \citep{DBLP:conf/iclr/IlharcoRWSHF23,yu2023language}. 

Due to the inherent correlations between delta parameters and post-training, significant efforts have been made to investigate the properties of delta parameters through various editing operations in recent years. For instance, studies like DARE \citep{yu2023language} and DELLA-Merging \citep{DBLP:journals/corr/abs-2406-11617} showed that models can achieve comparable performance with only a small fraction of delta parameters, highlighting their extreme redundancy. BitDelta \citep{DBLP:journals/corr/abs-2402-10193} demonstrated that delta parameters could be quantized to 1 bit with modest performance compromise. Twin-Merging \citep{DBLP:journals/corr/abs-2406-15479} and TIES-Merging \citep{DBLP:conf/nips/YadavTCRB23} discovered that most of the benefits of post-training can be retained after executing singular value decomposition and magnitude-based pruning on delta parameters. EXPO \citep{DBLP:journals/corr/abs-2404-16792} observed that cheaply extrapolating delta parameters with a suitable scaling factor can even enhance the performance. However, a comprehensive framework for systematically discussing delta parameter characteristics and theoretically explaining how different operations impact model performance remains lacking. 

In this work, we make a pioneering effort to provide a unified view of delta parameter editing in post-trained large-scale models. We formulate the editing operations of delta parameters based on Riemann sum approximation of the loss difference. By mathematically analyzing existing editing operations' loss change, we elucidate why certain operations result in competitive, decreased, or improved performance. Specifically, we verify that: 1) methods such as DARE and DELLA-Merging can well keep the approximation term to zero through the random drop and rescale processes, ensuring equal loss between the edited and post-trained models and achieving competitive performance; 2) techniques including BitDelta, Twin-Merging, and TIES-Merging often result in decreased performance, with a positive approximation term introduced by quantization, low-rank approximation, and magnitude-based pruning; 
3) EXPO-like methods, by extrapolating delta parameters, produce negative loss changes on alignment data, resulting in better-aligned models.
To validate our theoretical analysis, extensive experiments are conducted on large-scale visual models (ViT \citep{radford2021learning}) and language models (LLaMA 3 \citep{DBLP:journals/corr/abs-2407-21783}, and Mistral \citep{DBLP:journals/corr/abs-2310-06825}), and the results strongly support our analysis. 

Besides understanding existing delta parameter editing techniques in the proposed view, we further present several extensions to provide more general formats. 
Firstly, we introduce a factor to handle the dropped parameters in DARE, effectively expanding methods like DARE. 
Secondly, we extend the scope of quantification-based methods like BitDelta, identifying a broader area for effective quantification beyond reducing magnitude diversity to a single value. 
Finally, we identify that extrapolation is not the key to the success of EXPO-like methods. Instead, we should determine whether to use extrapolation or interpolation based on the direction of the approximation term.
Experimental results also demonstrate the effectiveness of the proposed extensions. 

\section{Related Work}
\label{section-2}

\subsection{Post-training of Large-Scale Models}
In recent years, with the rapid development of large-scale models, post-training has become an essential process for adapting the pre-trained backbone to a variety of tasks \citep{DBLP:journals/corr/abs-2402-02242,DBLP:journals/corr/abs-2002-06305,DBLP:journals/corr/abs-2303-18223}. Post-training realizes the adaptation via adjusting the pre-trained backbone's parameters through full fine-tuning \citep{DBLP:conf/iclr/DosovitskiyB0WZ21,DBLP:conf/iccv/LiuL00W0LG21,DBLP:conf/naacl/DevlinCLT19,radford2018improving} or parameter-efficient fine-tuning \citep{DBLP:conf/aaai/HeLZYW23,DBLP:conf/icml/HoulsbyGJMLGAG19,DBLP:conf/acl/LiL20,DBLP:conf/iclr/HuSWALWWC22,DBLP:journals/corr/abs-2403-14608} algorithms. It is straightforward to conclude that the effectiveness of post-training can be perfectly denoted by the delta parameters, which represent the difference between post-trained and pre-trained parameters \citep{DBLP:conf/iclr/IlharcoRWSHF23,yu2023language}. Given the close correlations between delta parameters and the post-training process, investigating the properties of delta parameters becomes particularly important. In this paper, we present a novel perspective to illustrate delta parameter characteristics of post-trained models. 

\subsection{Delta Parameter Editing for Post-Trained Models}
Existing delta parameter editing techniques can be generally categorized as three aspects according to their post-editing performance, including competitive, decreased, and improved performance.

\textbf{Delta Parameter Editing with Competitive Performance}.
DARE \citep{yu2023language} is a widely used approach to edit delta parameters without compromising the model performance. Technically, DARE can eliminate most (90\% or even 99\%) of the delta parameters with the random drop and rescale operations. Inspired by DARE, DELLA-Merging \citep{DBLP:journals/corr/abs-2406-11617} presented a magnitude-aware drop to replace the random drop for achieving better performance, which ranks delta parameters by their magnitude and assigns higher dropout probabilities to those with lower ranks (i.e., corresponding to lower magnitudes). \citet{yu2023language} and \citet{DBLP:journals/corr/abs-2406-11617} explained that DARE and DELLA-Merging can work because they are able to approximate the original embeddings based on only a small fraction of delta parameters, thus maintaining the model performance. 

\textbf{Delta Parameter Editing with Decreased Performance}.
BitDelta \citep{DBLP:journals/corr/abs-2402-10193} quantized delta parameters to only 1 bit according to the average magnitude scalar and sign bits. Twin-Merging \citep{DBLP:journals/corr/abs-2406-15479} applied singular value decomposition \citep{klema1980singular} on delta parameters to extract exclusive knowledge for each specific task. TIES-Merging \citep{DBLP:conf/nips/YadavTCRB23} retained delta parameters with the largest magnitudes for reducing redundancy. All the above methods yield slightly worse results after executing the corresponding quantization, low-rank approximation, or pruning operations.
 
\textbf{Delta Parameter Editing with Improved Performance}.
EXPO \citep{DBLP:journals/corr/abs-2404-16792} extrapolated delta parameters calculated by two relatively weaker models with an appropriate scaling factor to construct a stronger model, which can enhance the model performance. 

It can be concluded that current approaches utilizes distinct operations for editing delta parameter, lacking a comprehensive analysis of whether these editing operations are suitable and why different operations cause various influence on the model performance. In this work, we make the first attempt to introduce a unified view of delta parameter editing in post-training, which is supported both theoretically and empirically.

\section{Preliminaries}
\label{section-3}

\subsection{Notations}
\textbf{Delta Parameters During Post-Training}. Let $\bm{W}_{\text{PRE}} \in \mathbb{R}^{d \times k}$ denote the parameters of a pre-trained model, where $d$ and $k$ represent the output and input dimensions. A post-trained model with parameters $\bm{W}_{\text{POST}} \in \mathbb{R}^{d \times k}$ can be derived from the pre-trained backbone, yielding delta parameters $\Delta \bm{W} = \bm{W}_{\text{POST}} - \bm{W}_{\text{PRE}} \in \mathbb{R}^{d \times k}$. As delta parameters denote the alterations of parameters during the post-training process, investigating the characteristics of delta parameters can provide a deeper understanding of post-training.

\textbf{Delta Parameter Editing}. Let $\mathcal{F}$ represent the delta parameter editing function. The edited parameters $\Delta \widetilde{\bm{W}}_{\text{Edit}} = \mathcal{F}(\Delta \mathbf{W})$ is then combined with $\mathbf{W}_{\text{PRE}}$ to obtain the final edited parameter $\mathbf{W}_{\text{Edit}} = \mathbf{W}_{\text{PRE}} + \Delta \widetilde{\bm{W}}_{\text{Edit}}$. Existing delta parameter editing methods can be categorized into three types based on their effects on model performance, i.e., competitive, decreased, and improved performance. These methods employ various techniques including pruning, quantization, low-rank approximation, and extrapolation. Notable works in this field include DARE, BitDelta, Twin-Merging, TIES-Merging, and EXPO, which are investigated in this paper.


\subsection{A Unified View of Delta Parameter Editing}
In this work, we introduce a unified view of delta parameter editing during the post-training process based on Riemann sum approximation. 
Specifically, we represent the changes caused by existing editing methods by $\Delta \widetilde{\bm{W}}$ and aim to investigate their effects on performance via analyzing the loss difference. To better analyze the changes in loss, we introduce the Riemann sum approximation, which corresponds to the difference in loss made by the editing operation as follows,
\begin{equation}
    \label{equ:gradient_multiple_delta}
    \begin{aligned}
    \Delta \mathcal{L} &= \mathcal{L}(\bm{W}_{\text{POST}} + \Delta \widetilde{\bm{W}}) - \mathcal{L}(\bm{W}_{\text{POST}}) = \int_{0}^{1} \nabla \mathcal{L}(\bm{W}_{\text{POST}} + t \Delta \widetilde{\bm{W}}) \cdot \Delta \widetilde{\bm{W}} \, dt \\
    &\approx \frac{1}{C}\sum\limits_{c=0}^{C-1}\langle \nabla\mathcal{L}(\bm{W}_{\text{POST}} + \frac{c}{C} \Delta \widetilde{\bm{W}}), \Delta \widetilde{\bm{W}}\rangle = \frac{1}{C}\sum\limits_{c=0}^{C-1}\langle \nabla\mathcal{L}^c, \Delta \widetilde{\bm{W}}\rangle,
    \end{aligned}
\end{equation}
where $\mathcal{L}(\bm{W}):\mathbb{R}^{d \times k} \rightarrow \mathbb{R}$ denotes the loss function of a model with parameters $\bm{W} \in \mathbb{R}^{d \times k}$,
$\nabla \mathcal{L}(\bm{W})$ is the gradient of the loss function at $\bm{W}$, and $\langle \cdot, \cdot \rangle$ denotes the Frobenius inner product. $C$ denotes the number of subdivisions of the interval $[0, 1]$. This expansion provides a linear approximation of the loss function in the neighborhood of $\bm{W}_{\text{POST}}$, allowing the analysis of the impact of parameter changes on the model performance.
In most cases, the loss difference can reflect the influence on performance, with a positive value indicating deterioration, zero indicating stability, and a negative value indicating improvement. 
In \secref{section-4}, \secref{section-5}, and \secref{section-6}, we respectively discuss editing operations that cause competitive, decreased, and improved performance, and derive the format of these operations when organizing them into the proposed unified paradigm.

To validate our theoretical analysis and the proposed extensions, we conducted experiments on LLaMA-3-8B-Instruct~\citep{DBLP:journals/corr/abs-2407-21783}, Mistral-7B-Instruct-v0.3~\citep{DBLP:journals/corr/abs-2310-06825}, and ViT-B-32~\citep{radford2021learning}. We evaluate text models on 8 tasks: 25-shot ARC Challenge~\citep{clark2018think}, 5-shot GSM8K~\citep{cobbe2021training}, 10-shot HellaSwag~\citep{zellers2019hellaswag}, zero-shot HumanEval~\citep{chen2021evaluating}, zero-shot IFEval~\citep{zhou2023instructionfollowingevaluationlargelanguage}, 5-shot MMLU~\citep{hendrycks2020measuring}, zero-shot TruthfulQA~\citep{lin2021truthfulqa}, and 5-shot Winogrande~\citep{sakaguchi2021winogrande}, and evaluate vision models on 8 tasks: Cars~\citep{krause20133d}, DTD~\citep{cimpoi2014describing}, EuroSAT~\citep{helber2019eurosat}, GTSRB~\citep{stallkamp2011german}, MNIST~\citep{lecun2010mnist}, RESISC45~\citep{cheng2017remote}, SUN397~\citep{xiao2016sun}, and SVHN~\citep{netzer2011reading}.

\section{Unifying Editing Operations with Competitive Performance}
\label{section-4}

As a widely-used approach for delta parameter editing, DARE \citep{yu2023language} presents the random drop and rescale process to remove 90\% or even 99\% delta parameters without compromising the model performance. Following this line, many follow-up works have been proposed. For example, DELLA-Merging \citep{DBLP:journals/corr/abs-2406-11617} modifies the drop operation in DARE from random to magnitude-aware. In this section, we select DARE for analysis because it is the most representative method among those that can retain the original model performance after editing delta parameters.

\subsection{Express DARE with Approximation Term}
Mathematically, the editing process of delta parameters in DARE is denoted by
\begin{equation}
\begin{split}
    \label{equ:dare_computation}
    & \bm{W}_{\text{DARE}} = \bm{W}_{\text{POST}} + \Delta \widetilde{\bm{W}}_{\text{DARE}} =  \bm{W}_{\text{PRE}} + \Delta \bm{W} + \Delta \widetilde{\bm{W}}_{\text{DARE}} \\ =\bm{W}_{\text{PRE}} + 0 \cdot \bm{M} & \odot \Delta \bm{W} + \frac{1}{1-p} \cdot (1 - \bm{M}) \odot \Delta \bm{W} = \bm{W}_{\text{PRE}} + \frac{1}{1-p} \cdot (1 - \bm{M}) \odot \Delta \bm{W},
\end{split}
\end{equation}

where $p \in \mathbb{R}$ represents the drop rate and $\odot$ denotes the element-wise Hadamard product. $\bm{M} \sim \text{Bernoulli}(p, \Delta \bm{W}) \in \mathbb{R}^{d \times k}$ is a mask matrix sampled from Bernoulli distribution according to $p$, whose shape is identical to that of $\Delta \bm{W}$. From \equref{equ:dare_computation}, we can derive that 
\begin{equation}
    \label{equ:dare_introduced_delta}
    \Delta \widetilde{\bm{W}}_{\text{DARE}} = \frac{p - \bm{M}}{1 - p} \odot \Delta \bm{W}.
\end{equation}
Referring to \equref{equ:gradient_multiple_delta}, we obtain
\begin{equation}
\label{equ:dare_first_order_term}
\begin{split}
\Delta\mathcal{L}_{\text{DARE}} &\approx \frac{1}{C}\sum\limits_{c=0}^{C-1} \sum\limits_{i=1}^{d} \sum\limits_{j=1}^{k} \frac{p - M_{ij}}{1 - p} \cdot \Delta W_{ij} \cdot \nabla \mathcal{L}^c_{ij} \\ 
& = \frac{1}{C}\sum\limits_{c=0}^{C-1} \left(\frac{p}{1-p} \cdot  \sum\limits_{M_{ij}=0} \Delta W_{ij} \cdot \nabla \mathcal{L}^c_{ij} -  \sum\limits_{M_{ij}=1} \Delta W_{ij} \cdot \nabla \mathcal{L}^c_{ij}\right).
\end{split}
\end{equation}
Due to the vast number of parameters in large-scale models, we can use the Law of Large Numbers to approximate the summations by their expected values. Additionally, because of the randomness of the drop operation in DARE, $M_{ij}$ and $\Delta W_{ij} \cdot \nabla \mathcal{L}^c_{ij}$ can be considered approximately independent random variables. It is straightforward to deduce that
\begin{equation}
\label{equ:random_drop_expectation}
\begin{split}
\sum\limits_{M_{ij}=0} \Delta W_{ij} \cdot \nabla \mathcal{L}^c_{ij} &\approx  (1 - p) \cdot \sum\limits_{i=1}^{d} \sum\limits_{j=1}^{k} \Delta W_{ij} \cdot \nabla \mathcal{L}^c_{ij}, \\
\sum\limits_{M_{ij}=1} \Delta W_{ij} \cdot \nabla \mathcal{L}^c_{ij} &\approx p \cdot \sum\limits_{i=1}^{d} \sum\limits_{j=1}^{k} \Delta W_{ij} \cdot \nabla \mathcal{L}^c_{ij}.
\end{split}
\end{equation}
Substituting \equref{equ:random_drop_expectation} into \equref{equ:dare_first_order_term}, we derive
\begin{equation}
\label{equ:final_dare_first_order_term}
\begin{split}
\Delta\mathcal{L}_{\text{DARE}} \approx  (\frac{p}{1-p} \cdot (1-p) - p) \cdot \frac{1}{C}\sum\limits_{c=0}^{C-1}\sum\limits_{i=1}^{d} \sum\limits_{j=1}^{k} \Delta W_{ij} \cdot \nabla \mathcal{L}^c_{ij} = 0.
\end{split}
\end{equation}

To this end, we can conclude that after editing delta parameters with DARE, the loss $\mathcal{L}(\bm{W}_{\text{DARE}})$ is approximately equal to $\mathcal{L}(\bm{W}_{\text{POST}})$, independent of the specific dataset, explaining why DARE can achieve competitive performance even when most delta parameters are eliminated.

To verify the above analysis, we used the DARE method to construct models on LLaMA3-8B-Instruct and computed the approximation term on the GSM8K dataset. The results are shown in Figure~\ref{fig:dare_gradient_delta}. We use the drop-only (w/o rescale) as the reference. As can be seen, models with DARE constructed consistently achieved lower average loss, and with a smaller drop rate, the approximation term calculated across different parts of the model remained relatively small. This validates our theoretical derivation above.


\subsection{Extension of DARE}
We further present a more general format of delta parameter editing operations that can achieve competitive performance. In particular, instead of dropping delta parameters, we introduce a term $k$ to adjust them and rescale the remaining ones with $(1 - k \cdot p)/(1-p)$. Similar to the deduction in \equref{equ:dare_computation} to  \equref{equ:final_dare_first_order_term}, we obtain

\begin{equation}
\label{equ:dare_extension}
\notag
\begin{split}
    \bm{W}_{\text{COMP}} = &\bm{W}_{\text{PRE}} + \Delta \bm{W} + \Delta \widetilde{\bm{W}}_{\text{COMP}} = \bm{W}_{\text{PRE}} + k \cdot \bm{M} \odot \Delta \bm{W} + \frac{1 - k \cdot p}{1-p} \cdot (1 - \bm{M}) \odot \Delta \bm{W}, \\
    \Delta\widetilde{\bm{W}}_{\text{COMP}} \,&=\, \frac{(k-1)(\bm{M} - p)}{1 - p} \odot \Delta \bm{W}, \\
    \Delta\mathcal{L} \,&\approx\, \frac{1}{C} \sum\limits_{c=0}^{C-1} \left( \frac{p \cdot (1 - k)}{1-p} \cdot  \sum\limits_{M_{ij}=0} \Delta W_{ij} \cdot \nabla \mathcal{L}^c_{ij} + (k-1) \cdot \sum\limits_{M_{ij}=1} \Delta W_{ij} \cdot \nabla \mathcal{L}^c_{ij} \right)\\
    &\approx \left(\frac{p \cdot (1 - k)}{1-p} \cdot  (1-p) + (k - 1) \cdot p\right) \cdot \frac{1}{C} \sum\limits_{c=0}^{C-1} \sum_{i=1}^{d} \sum_{j=1}^{k} \Delta W_{ij} \cdot \nabla \mathcal{L}^c_{ij} = 0.
\end{split}
\end{equation}
\begin{figure}[htbp]
  \centering
  \includegraphics[width=1.0\columnwidth]{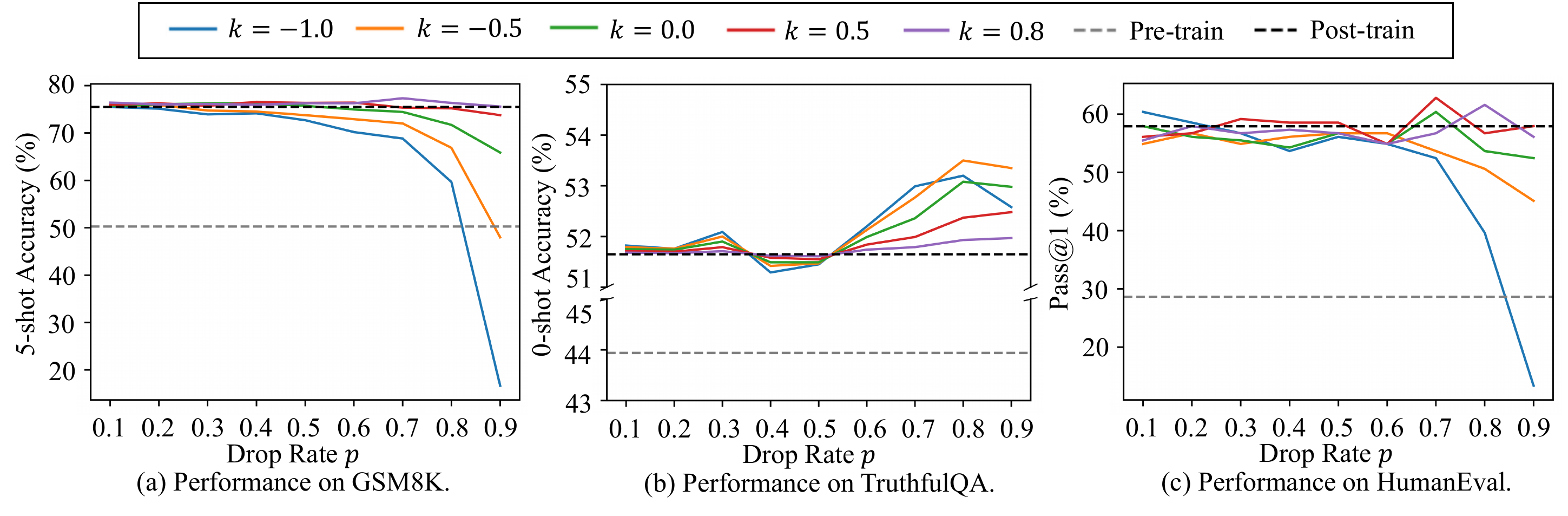} 
  \caption{The performance of LLaMA3-8B-Instruct on the GSM8K, TruthfulQA, and HumanEval datasets under varying $p$ and $k$.}
  \vspace{-10px}\label{fig:dare_main_performance}
\end{figure}

  
  
  

It has been verified that $\Delta\mathcal{L}$ is approximately 0, which indicates the validity of the proposed format. Note that in DARE, the drop operation can be realized by setting $k$ to 0. Thus, our format is an extension of DARE with broader settings of $k$.

\begin{figure}[htbp]
  \centering
  \includegraphics[width=1.0\columnwidth]{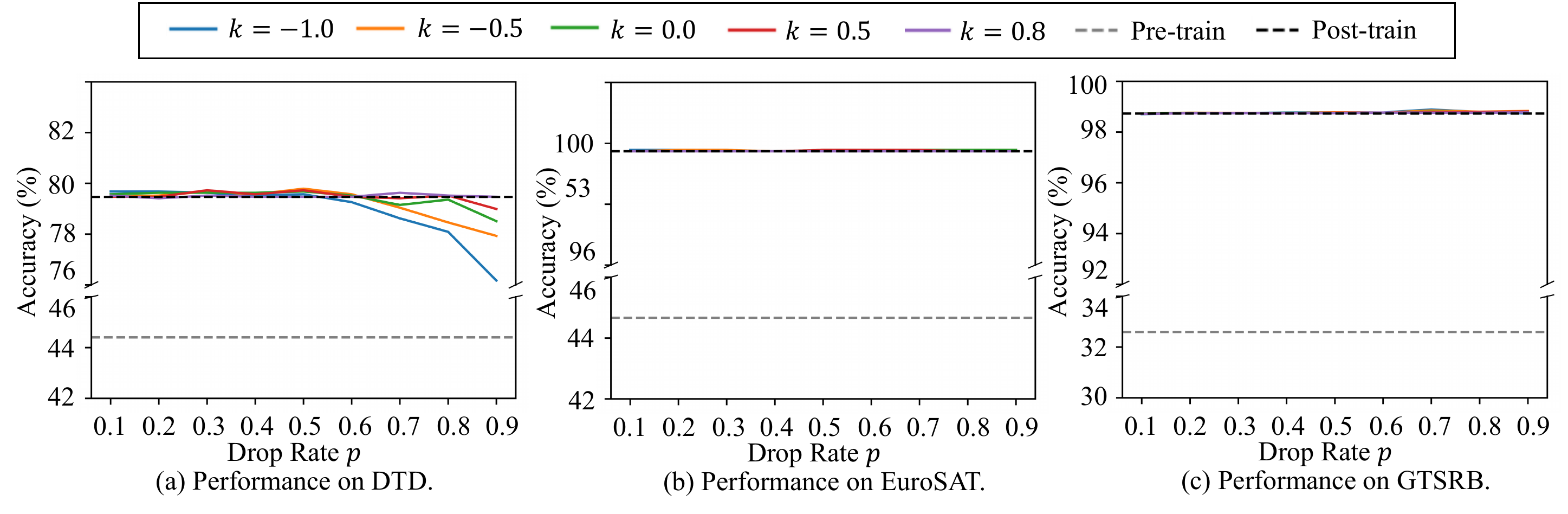} 
  \caption{The performance of ViT-B-32 on the DTD, EuroSAT, and GTSRB datasets under varying $p$ and $k$.}
  \vspace{-20px}\label{fig:dare_main_performance2}
\end{figure}

We conducted validation experiments for the extension of DARE across a wide range of models and tasks. The representative results for LLaMA3-8B-Instruct and ViT-B-32 are shown in~\ref{fig:dare_main_performance} and Figure~\ref{fig:dare_main_performance2}, while the complete results for all models and tasks are presented in the Appendix~\ref{appendix:dare}.
Specifically, on threee representative text datasets—GSM8K, TruthfulQA, and HumanEval, when both the rescale rate $k$ and sign change rate $kp$ are small (e.g., less than 0.5), the performance of our adjusted model is very close to that of the original post-trained model and significantly outperforms the pre-trained model. 
Regarding the weight scalar $k$ introduced in our extension, we observed that, compared to the setting where $k=0$ (which reverts to the original DARE configuration), using $k\neq 0$ generally yields competitive performance across different datasets. This demonstrates the effectiveness of our extension.
For the ViT model, the results on the DTD, EuroSAT, and GTSRB datasets are more consistent with our expectations. Regardless of the rescale and sign change rates, the performance of the adjusted model is almost identical to that of the original post-trained model.

Interestingly, when $k < 0$, indicating that the delta parameters are flipped in sign, the constructed model still achieves competitive performance to the post-trained model. This challenges the prior assumption that the sign of delta parameters is critical for performance~\citep{DBLP:conf/nips/YadavTCRB23,DBLP:journals/corr/abs-2402-10193}, suggesting that what truly matters during post-training is not the specific directional adjustments of individual parameters, but rather a more collective behavior of the entire delta parameters.

\subsection{Further Discussions on DARE}

\citet{yu2023language} and \citet{DBLP:journals/corr/abs-2406-11617} claim that DARE and DELLA-Merging are effective because the random drop of delta parameters ensures an approximation of the original embeddings, thereby preserving model performance. However, according to \eqref{equ:random_drop_expectation}, we argue that random drop of delta parameters is a sufficient but not necessary condition for maintaining model performance. Furthermore, we contend that ensuring randomness in the element-wise product of delta parameters and approximation term is the necessary and sufficient condition.
\begin{wraptable}{r}{0.6\textwidth}
\centering
\begin{tabular}{cccc}
\toprule
k   & Random & Biased $\Delta W$ & Biased $\Delta W\cdot\nabla L$ \\ 
\midrule
0.5 & 76.35  & 74.15               & 0.0                  \\
0.7 & 75.89  & 75.36               & 0.0                  \\
0.9 & 76.19  & 76.04               & 26.76                \\
1.1 & 75.89  & 75.59               & 0.15                 \\
1.3 & 75.36  & 74.91               & 0.0                  \\
1.5  & 75.59  & 74.83               & 0.0 \\ 
\bottomrule
\end{tabular}
\caption{Validation of the discussion on DARE. The leftmost column shows the random drop in DARE. The middle column illustrates the approach of multiplying all negative delta parameters by $k$ and all positive delta parameters by $ \frac{1 - k \cdot p}{1 - p} $. The rightmost column demonstrates the method of first calculating the product of delta parameters and gradients, and then multiplying all negative products by $ k $ and all positive products by $ \frac{1 - k \cdot p}{1 - p} $.}
\vspace{-5px}
\label{tab:dare_discussion}
\end{wraptable}

To verify the above analysis, we conduct two experiments on GSM8K dataset.
First, we disrupt the randomness of the delta parameter drop operation by multiplying all negative delta parameters by $k$ and all positive delta parameters by $(1 - k \cdot p)/(1-p)$. The results are shown in the middle column of Table~\ref{tab:dare_discussion}, illustrating that the model performance remains intact. This validates that the randomness of the delta parameter dropout operation is a sufficient but not necessary condition for maintaining model performance.
Furthermore, we disrupt the randomness of the dropout operation on the approximation term by multiplying all negative products by $k$ and all positive products by $(1 - k \cdot p)/(1-p)$. The results, as depicted in the rightmost of Table~\ref{tab:dare_discussion}, show a significant decline in model performance. This validates that the randomness of the dropout operation on the product of delta parameters and approximation term is a necessary and sufficient condition for maintaining model performance.


\begin{figure}[htbp]
  \centering
  \includegraphics[width=0.95\columnwidth]{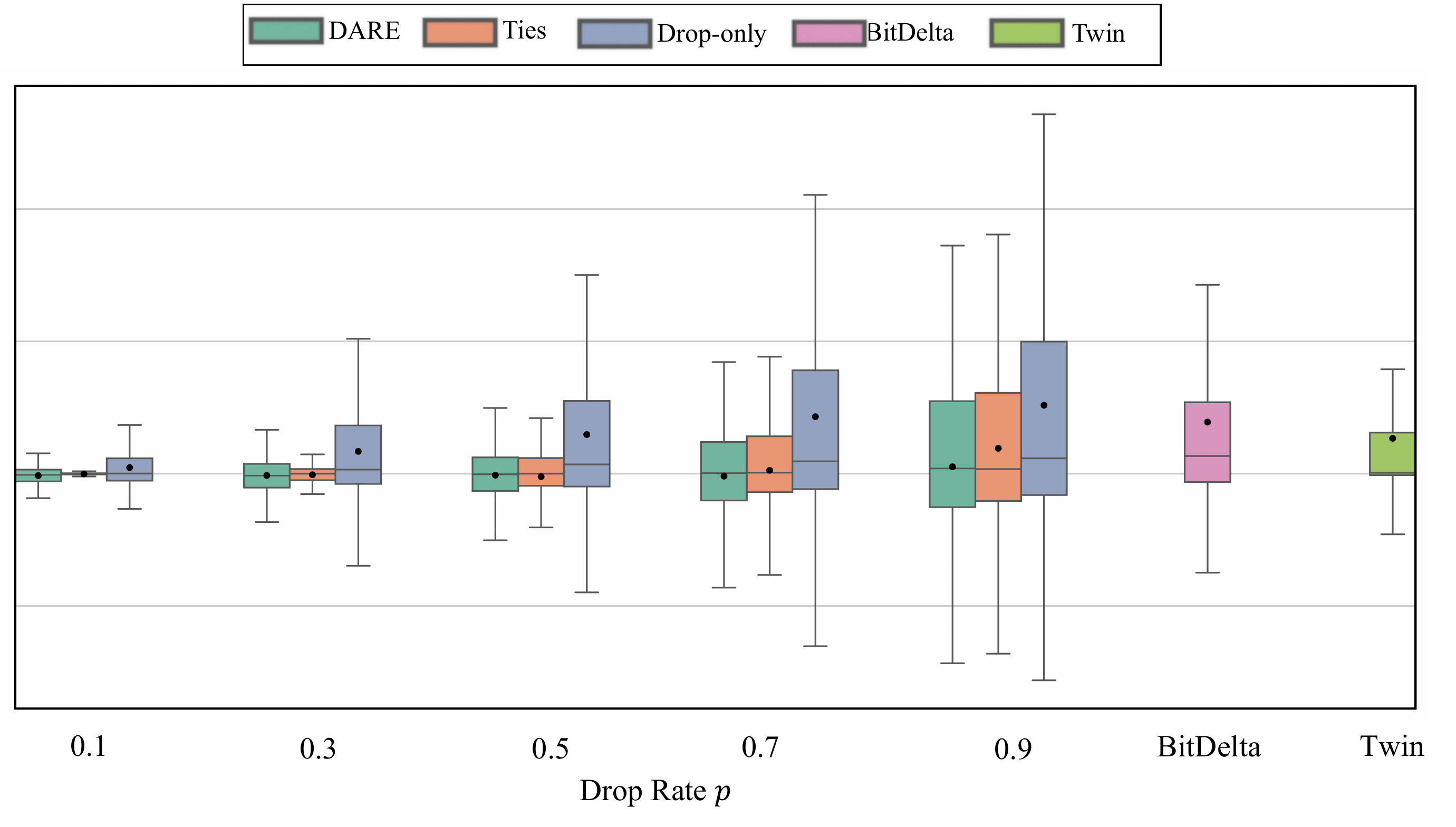} 
  \caption{Validation of our theoretical derivation of DARE, BitDelta, Twin-Merge(sparsity rate=0.9), and Ties-Merge.}
  \label{fig:dare_gradient_delta}
\end{figure}

\section{Unifying Editing Operations with Decreased Performance}
\label{section-5}
This section discusses three delta parameter editing operations that incur reduced results, including quantization, low-rank approximation, and pruning. We respectively choose BitDelta \citep{DBLP:journals/corr/abs-2402-10193}, Twin-Merging \citep{DBLP:journals/corr/abs-2406-15479}, and TIES-Merging \citep{DBLP:conf/nips/YadavTCRB23} as typical works.

\subsection{Express BitDelta with Approximation Term}
BitDelta quantizes delta parameters down to 1 bit, utilizing the sign bit matrix and a high-precision scalar, where the latter is initially computed by the average magnitude of delta parameters. Specifically, BitDelta can be represented by
\begin{equation}
\begin{split}
    \label{equ:bitdelta_computation}
    & \bm{W}_{\text{BitDelta}} = \bm{W}_{\text{POST}} + \Delta \widetilde{\bm{W}}_{\text{BitDelta}} =  \bm{W}_{\text{PRE}} + \Delta \bm{W} + \Delta \widetilde{\bm{W}}_{\text{BitDelta}} \\ 
    =\bm{W}_{\text{PRE}} + & \frac{1}{d \cdot k} \sum\limits_{i=1}^{d} \sum\limits_{j=1}^{k} | \Delta W_{ij} | \cdot \text{Sign}(\Delta \bm{W}) = \bm{W}_{\text{PRE}} + \text{AVG}(|\Delta \bm{W}|) \cdot \text{Sign}(\Delta \bm{W}),
\end{split}
\end{equation}
where $|\cdot|$ denotes the operation of taking magnitudes. $\text{AVG}(|\Delta \bm{W}|)$ represents the average magnitude of $\Delta \bm{W}$. Since $\Delta \bm{W} = |\Delta \bm{W}| \odot \text{Sign}(\Delta \bm{W})$, based on \equref{equ:bitdelta_computation}, we can further obtain
\begin{equation}
    \label{equ:bitdelta_introduced_delta}
    \Delta \widetilde{\bm{W}}_{\text{BitDelta}} = (\text{AVG}(|\Delta \bm{W}|) - |\Delta \bm{W}|) \odot \text{Sign}(\Delta \bm{W}).
\end{equation}
Based on \equref{equ:gradient_multiple_delta}, we get
\begin{equation}
\label{equ:bitdelta_first_order_term}
\begin{split}
\Delta\mathcal{L}_{\text{BitDelta}} \approx \frac{1}{C} \sum\limits_{c=0}^{C-1} \sum\limits_{i=1}^{d} \sum\limits_{j=1}^{k}  (\text{AVG}(|\Delta \bm{W}|) - |\Delta W_{ij}|) \cdot \text{Sign}(\Delta W_{ij}) \cdot \nabla \mathcal{L}^c_{ij}.
\end{split}
\end{equation}
Though $\sum\limits_{i=1}^{d} \sum\limits_{j=1}^{k}  ((\text{AVG}(|\Delta \bm{W}|) - |\Delta W_{ij}|) = d \cdot k \cdot \text{AVG}(|\Delta \bm{W}|) - \sum\limits_{i=1}^{d} \sum\limits_{j=1}^{k} |\Delta W_{ij}| = 0$, it is hard to conclude that \equref{equ:bitdelta_first_order_term} equals 0 due to the multiplication of $\text{Sign}(\Delta W_{ij}) \cdot \nabla \mathcal{L}^c_{ij}$. Based on the approximation of the loss in Figure~\ref{fig:dare_gradient_delta}, it can be observed that the loss of BitDelta on GSM8K is greater than 0, which is consistent with its performance degradation on GSM8K compared to post-trained model.

\subsection{Express Twin-Merging and TIES-Merging with Approximation Term}
Twin-Merging employs singular value decomposition on delta parameters to derive task-specific knowledge. TIES-Merging preserves delta parameters with the highest magnitudes to minimize redundancy. Their computation processes are
\begin{equation}
\begin{split}
    \label{equ:twin_ties_merging_computation}
    \bm{W}_{\text{Twin}} & = \bm{W}_{\text{POST}} + \Delta \widetilde{\bm{W}}_{\text{Twin}} =  \bm{W}_{\text{PRE}} + \Delta \bm{W} + \Delta \widetilde{\bm{W}}_{\text{Twin}} = \bm{W}_{\text{PRE}} + \bm{U}_r \bm{\Sigma}_r \bm{V}_r^T, \\
    \bm{W}_{\text{TIES}} & = \bm{W}_{\text{POST}} + \Delta \widetilde{\bm{W}}_{\text{TIES}} =  \bm{W}_{\text{PRE}} + \Delta \bm{W} + \Delta \widetilde{\bm{W}}_{\text{TIES}} = \bm{W}_{\text{PRE}} + \bm{M} \odot \Delta \bm{W},
\end{split}
\end{equation}
where rank $r \leq \min(d, k)$ denotes the number of linearly independent columns (or rows) in $\Delta \bm{W} = \bm{U} \bm{\Sigma} \bm{V}^T$. $\bm{U}_r \in \mathbb{R}^{d \times r}$ consists of the first $r$ columns of $\bm{U}$ (whose columns are the left singular vectors of $\Delta \bm{W}$). $\bm{\Sigma}_r$ is the $r \times r$ diagonal matrix containing the top $r$ singular values. $\bm{V}_r \in \mathbb{R}^{k \times r}$ includes the first $r$ columns of $\bm{V}$ (whose columns are the right singular vectors of $\Delta \bm{W}$). $\bm{M} \in \mathbb{R}^{d \times k}$ is a binary mask matrix where an entry of 1 indicates that the corresponding delta parameter is among the top-$n$ percent in magnitude. $n$ is the proportion of delta parameters to be retained.
According to \equref{equ:twin_ties_merging_computation}, we derive 
\begin{equation}
\begin{split}
    \label{equ:twin_ties_merging_introduced_delta}
    \Delta \widetilde{\bm{W}}_{\text{Twin}} & = \bm{U}_r \bm{\Sigma}_r \bm{V}_r^T - \Delta \bm{W}, \\
    \Delta \widetilde{\bm{W}}_{\text{TIES}} & = \bm{M} \odot \Delta \bm{W} - \Delta \bm{W} = - \neg \bm{M} \odot \Delta \bm{W},
\end{split}
\end{equation}
where $\neg \bm{M}$ is the element-wise NOT operation. Based on \equref{equ:gradient_multiple_delta}, we get
\begin{equation}
\label{equ:twin_ties_merging_first_order_term}
\begin{split}
\Delta\mathcal{L}_{\text{Twin}} &\approx \frac{1}{C} \sum\limits_{c=0}^{C-1} \sum\limits_{i=1}^{d} \sum\limits_{j=1}^{k} ( {\bm{U}_r \bm{\Sigma}_r \bm{V}_r^T}_{ij} - \Delta W_{ij}) \cdot \nabla \mathcal{L}^c_{ij}, \\
\Delta\mathcal{L}_{\text{TIES}} &\approx  - \frac{1}{C} \sum\limits_{c=0}^{C-1}  \sum\limits_{i=1}^{d} \sum\limits_{j=1}^{k}  \neg M_{ij} \cdot \Delta W_{ij} \cdot \nabla \mathcal{L}^c_{ij}.
\end{split}
\end{equation}
We exploit the value of the approximation term through experiments.
Models were constructed using LLaMA3-8B-Instruct, and the approximation term was calculated on the GSM8K dataset. As shown in Figure~\ref{fig:dare_gradient_delta}, for TIES-Merging, when the drop rate is relatively low, the approximation term is also lower. However, as the drop rate increases to a certain level (e.g., 0.9), the performance begins to degrade compared to DARE, which is consistent with the observations in DARE. For Twin-Merging, the approximation loss is greater than zero, which aligns with the observed performance degradation on the GSM8K dataset.

\subsection{Extension of BitDelta}


We also extend the applicability of BitDelta by offering a more general form. Firstly, in addition to selecting the signs of delta parameters, we hypothesize that the effectiveness of BitDelta may stem from its choice of a holistic statistic that reflects the properties of the delta parameters.
Specifically, BitDelta utilizes the average magnitude of delta parameters to achieve the best approximation error in the $L_2$ norm.
To validate this, we conduct an experiment where we alter the holistic statistic selected by BitDelta, introducing varying degrees of noise to the average value.
As illustrated in the "Degenerate" line of Figure~\ref{fig:bitdelta_extension}, using the true average magnitude of the delta parameters yields nearly optimal performance on GSM8K, TruthfulQA, and HumanEval.
The performance changes along the degenerate line are quite steep, and slight modifications to this average value may result in a degradation of model performance.


  

\begin{figure}[htbp]
  \centering
  \includegraphics[width=0.8\columnwidth]{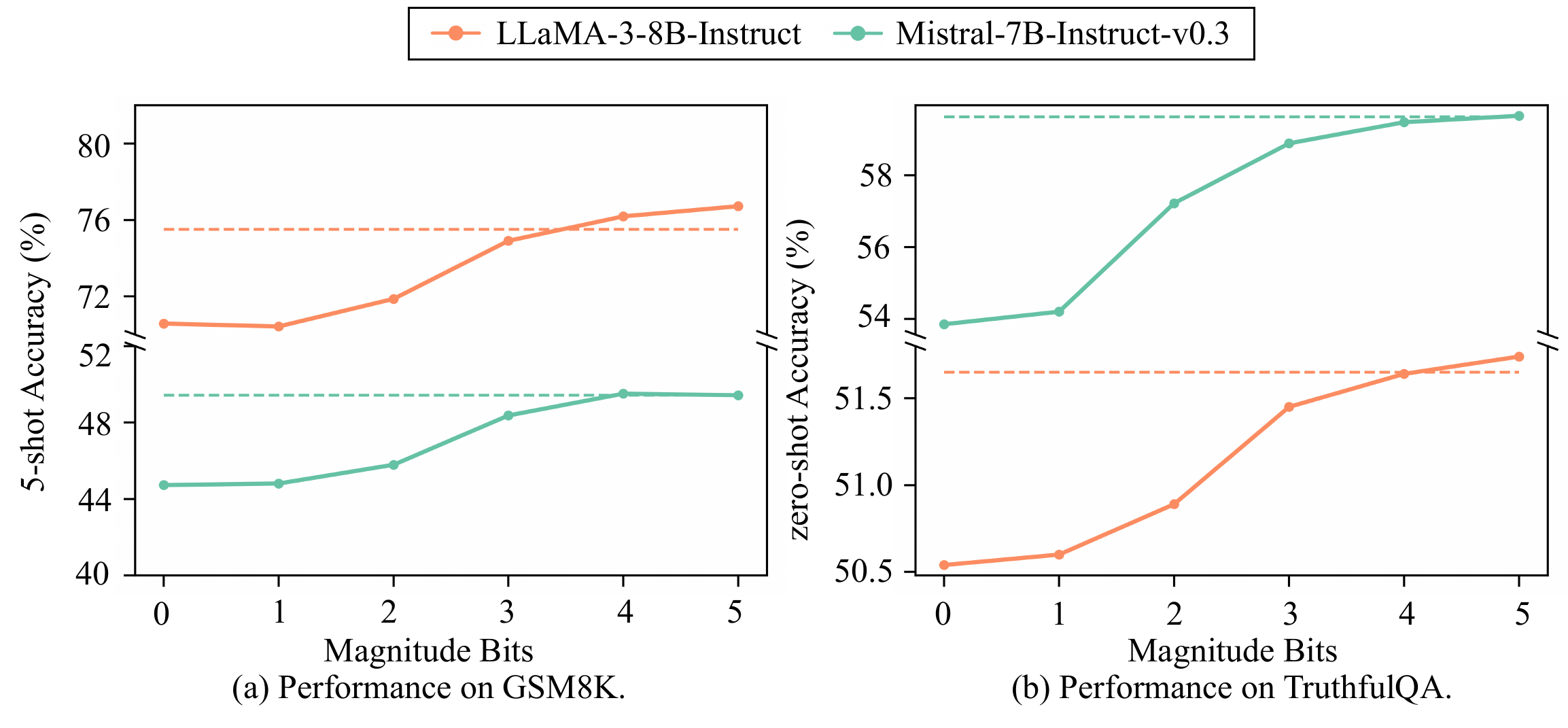}

  \caption{Effectiveness of increasing the number of bits in BitDelta. The left subplot shows the performance of LLaMA3-8B-Instruct and Mistral-7B-Instruct-v0.3 on the GSM8K dataset as the number of bits increases. The right subplot shows the performance on the TruthfulQA dataset. In each subplot, we use the dashed line to represent the performance of the original post-trained model.}
  \label{fig:bitdelta_extension}
\end{figure}

Secondly, instead of using a single value, we sample delta parameter magnitude matrices from both standard normal and uniform distributions, with the average magnitude serving as the mean. The experimental results, as depicted in Figure~\ref{fig:bitdelta_extension}, demonstrate that even when these parameters are randomly sampled from distributions, the model performance remains on par with a statistic value used in BitDelta. This further underscores the significance of selecting an appropriate holistic statistic for the delta parameters.

Finally, while preserving the relative magnitude relationships of delta parameters, we enhance the effectiveness of BitDelta by employing multiple bits.
Specifically, we divide the delta parameters into $M$ blocks based on their magnitude, from smallest to largest. 
Each block is then represented by the average value of the delta parameters within that block. When $M=1$, this approach corresponds to BitDelta, and when $M$ equals the total number of parameters in the model, it degenerates to the original post-trained model. The number of bits used is given by $\log_2 M $.
As shown in Figure~\ref{fig:bitdelta_extension}, increasing the number of bits significantly improves the model performance. When the number of bits is 4, the performance already surpasses that of the original post-trained model. This again highlights the redundancy in the delta parameters and demonstrates the potential for further advancements by expanding the bit representation in BitDelta.

\section{Unifying Editing Operations with Improved Performance}
\begin{figure}[htbp]
  \centering
  \includegraphics[width=1.0\columnwidth]{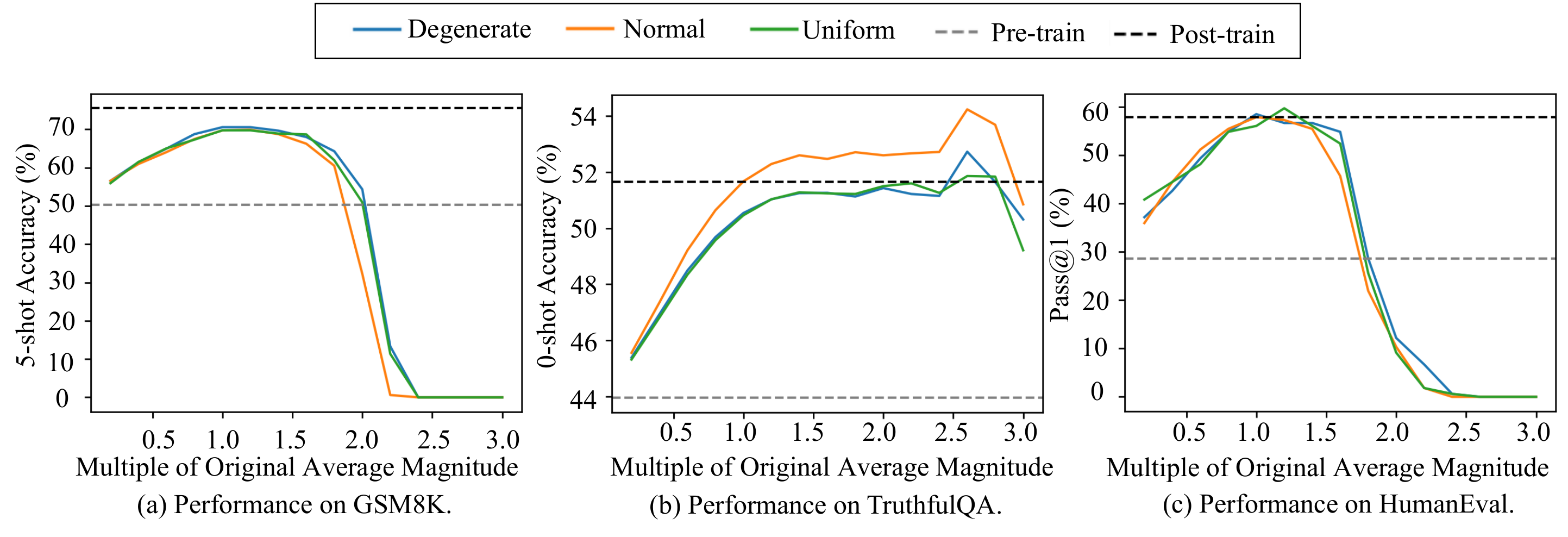}

  \caption{Validation of the extension of BitDelta. The degenerate curve at 1.0 represents the original BitDelta. The full results on 8 datasets are shown in Figure~\ref{fig:app_bitdelta_llama_all}.}
  \label{fig:bitdelta_extension}
\end{figure}

\label{section-6}
EXPO \citep{DBLP:journals/corr/abs-2404-16792} is a recent method to extrapolate delta parameters, which can boost LLMs' alignment. This section chooses EXPO as the representative approach for illustration.

\subsection{Express EXPO with Approximation Term}
Technically, EXPO first computes delta parameters between an aligned model and its initial fine-tuning checkpoints, and then extrapolates delta parameters with a suitable scaling factor for obtaining a better-aligned model. The calculation procedure is
\begin{equation}
    \label{equ:expo_computation}
    \bm{W}_{\text{EXPO}} = \bm{W}_{\text{POST}} + \Delta \widetilde{\bm{W}}_{\text{EXPO}} =  \bm{W}_{\text{PRE}} + \Delta \bm{W} + \Delta \widetilde{\bm{W}}_{\text{EXPO}} =\bm{W}_{\text{PRE}} + \Delta \bm{W} + \alpha \Delta \bm{W},
\end{equation}
where $\alpha$ controls the extrapolation length. Based on \equref{equ:expo_computation}, we derive
\begin{equation}
    \label{equ:expo_introduced_delta}
    \Delta \widetilde{\bm{W}}_{\text{EXPO}} = \alpha \Delta \bm{W}.
\end{equation}
Referring to \equref{equ:gradient_multiple_delta}, we obtain
\begin{equation}
\label{equ:expo_first_order_term}
\Delta\mathcal{L}_{\text{EXPO}} \approx \frac{\alpha}{C} \cdot  \sum\limits_{c=0}^{C-1}   \sum\limits_{i=1}^{d} \sum\limits_{j=1}^{k} \Delta W_{ij} \cdot \nabla \mathcal{L}^c_{ij}.
\end{equation}


\begin{wrapfigure}[19]{r}{0.55\columnwidth}
  \centering
  \vspace{-10px}
  \includegraphics[width=0.55\columnwidth]{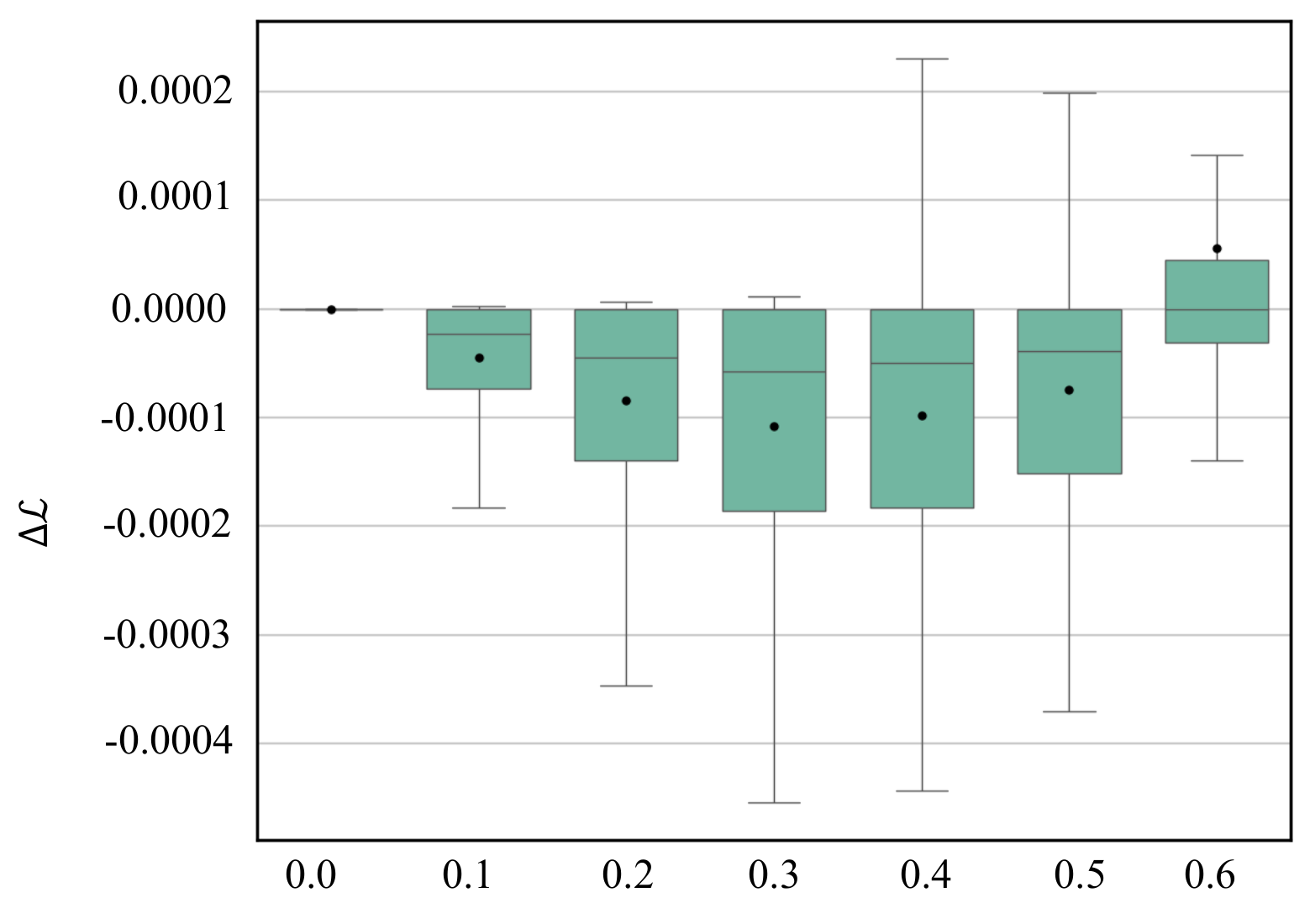} 
  \vspace{-20pt}
  \caption{Validation of our theoretical analysis of EXPO. we can observe that the approximation term first decreases and then increases as alpha changes, indicating that optimal performance is achieved at the trough.}
  \label{fig:expo_gradient_delta_04_dpo}
\end{wrapfigure}

An intuitive explanation for the improvements that EXPO achieves is that the DPO/RLHF training process of these models is suboptimal, which leads to the direction of loss reduction (the negative gradient) still aligning with the direction of the delta parameters, causing \equref{equ:expo_first_order_term} to be negative. Consequently, the loss of the edited model on alignment dataset is lower than that of the original post-training model, resulting in enhanced performance on alignment benchmarks.

We validated the aforementioned hypothesis by conducting experiments on Zephyr-7B-DPO-Full (trained by EXPO). We calculated the gradient of the models using DPO loss on UltraFeedback~\citep{cui2024ultrafeedback}. As shown in Figure~\ref{fig:expo_gradient_delta_04_dpo}, when $\alpha$ is relatively small, the value of the loss approximation term gradually decreases, reflecting that the model is indeed suboptimal. Moving further in this direction decreases the loss and improves performance accordingly. However, as $\alpha$ increases, the loss term gradually increases until it exceeds zero, which is consistent with the observation in EXPO that there is an optimal value for $\alpha$.

\subsection{Futher Discussions on EXPO}

\begin{wrapfigure}[20]{r}{0.55\columnwidth}
  \centering
  \vspace{-10px}\includegraphics[width=0.55\columnwidth]{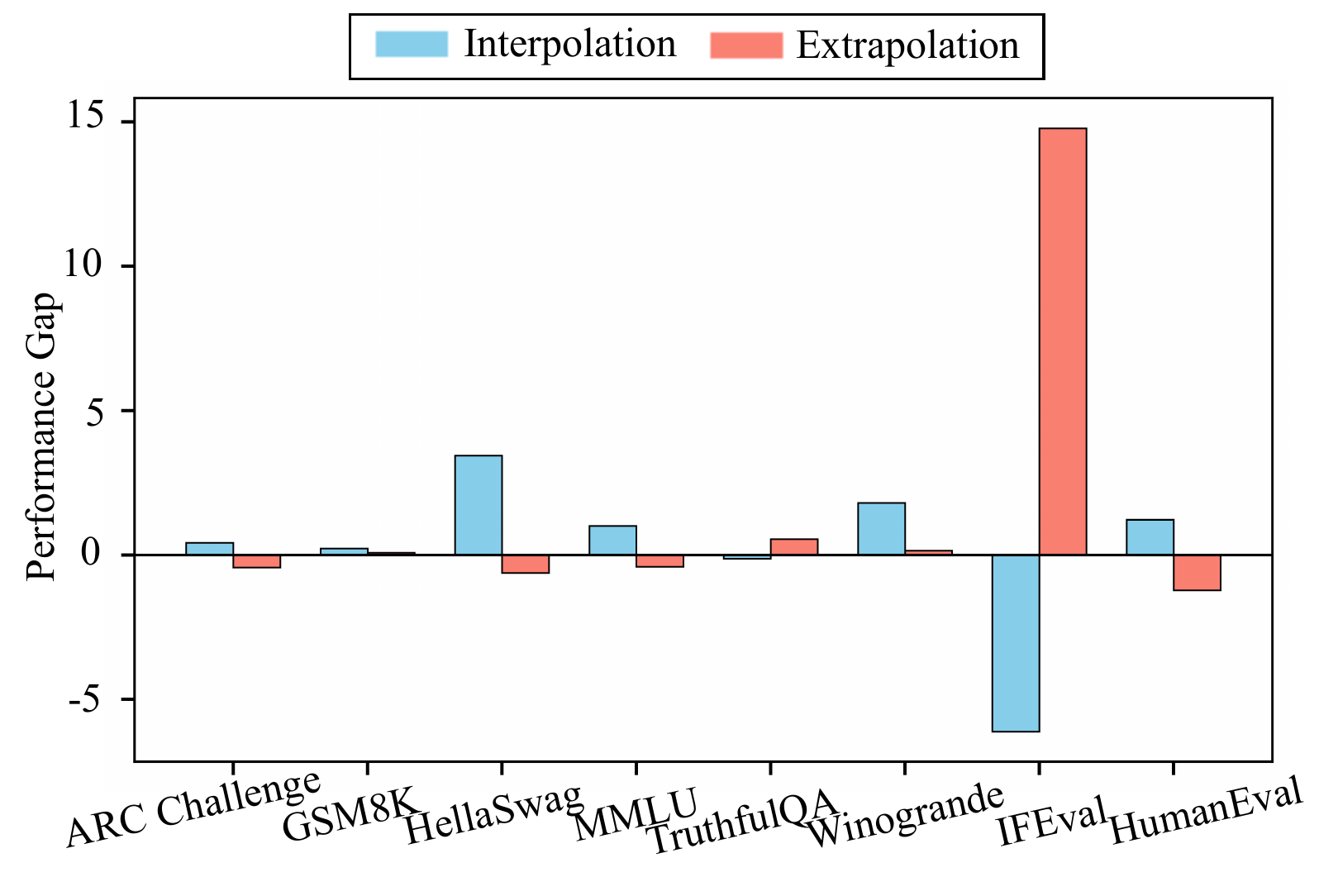} 
  \vspace{-10px}
  \caption{Comparison of Extrapolation and Interpolation Performance on LLaMA3-8B-Instruct. The performance gap represents the difference between the model's performance after extrapolation or interpolation and the original performance. }
  \label{fig:expo_discuss}
\end{wrapfigure}
EXPO claims that extrapolating delta parameters leads to better models. However, based on the derivation in \equref{equ:expo_first_order_term}, we believe that whether to use extrapolation or interpolation primarily depends on the direction of the gradient, which is influenced by the specific data. Specifically, for LLaMA3-8B-Instruct, we uniformly selected $\alpha$ in the range of -1.0 to 1.0 at intervals of 0.1, performing both interpolation and extrapolation of the model's delta parameters. As show in Figure~\ref{fig:expo_discuss}, on most datasets, interpolation outperformed extrapolation, except for the IFEval dataset, where extrapolation significantly improved performance. This confirms that whether to interpolate or extrapolate is not a fixed formula but depends on the specific data.We also conducted experiments on Qwen2-7B and Mistral-7B(shown in Section~\ref{appendix:expo_discuss}), and the results indicate that even for the same task, whether extrapolation or interpolation is required can vary across different models.

\section{Conclusion and Discussions}
\label{section-7}

Post-training is a core step in the training of large models. In recent years, significant efforts have been directed towards editing the delta parameters of post-training to achieve improvements in either performance or efficiency. However, while previous work has shown some effectiveness, the complexity of large model parameters has led to a fragmented understanding of delta parameter editing, with different studies focusing on different aspects of its effectiveness, lacking a unified perspective.

In this paper, we provide a unified perspective on the previous work related to post-training delta parameter editing using Riemann sum approximation. 
We find that the changes in model capability after altering the delta parameters can be analyzed through the loss differences approximated using Riemann sums.
By analyzing this approximation term, we can infer the reasons why the existing delta parameter editing methods lead to maintained, improved, or reduced model performance.


Our work offers a concise, unified, and powerful explanation for many previous work in the field of post-training delta parameter editing. We validate our hypothesis through numerical experiments. From our conclusions, several potential applications emerge for future work in this direction:
(1) Model Quantization: By finding an edit that sets the approximation term to zero while using lower precision, we can achieve nearly lossless compression of the model.
(2) Model Enhancement: By analyzing the approximation term, we might be able to find ways to enhance the model's capabilities without additional training data.
(3) Post-training Mechanism Analysis: Since the model's capability remains almost unchanged when the approximation term is zero, we can construct more concise post-training delta parameters.
This simplifies the parameter changes during the post-training phase, enabling a more effective analysis of the parameter mechanisms in this stage.

Additionally, our work highlights a critical observation: the analysis of parameter changes during the post-training phase should not be limited to specific parameters, such as knowledge neurons, but should consider the overall distribution of parameters. This is because the key constraint of the approximation term being zero does not depend on the changes in a specific parameter during post-training but requires a comprehensive consideration of all parameter deltas. This suggests that trying to infer the impact on the global model parameters from changes in a single or a few local parameters is likely futile.


\bibliography{reference}

\begin{thebibliography}{46}
\providecommand{\natexlab}[1]{#1}
\providecommand{\url}[1]{\texttt{#1}}
\expandafter\ifx\csname urlstyle\endcsname\relax
  \providecommand{\doi}[1]{doi: #1}\else
  \providecommand{\doi}{doi: \begingroup \urlstyle{rm}\Url}\fi

\bibitem[Chen et~al.(2021)Chen, Tworek, Jun, Yuan, Pinto, Kaplan, Edwards, Burda, Joseph, Brockman, et~al.]{chen2021evaluating}
Mark Chen, Jerry Tworek, Heewoo Jun, Qiming Yuan, Henrique Ponde De~Oliveira Pinto, Jared Kaplan, Harri Edwards, Yuri Burda, Nicholas Joseph, Greg Brockman, et~al.
\newblock Evaluating large language models trained on code.
\newblock \emph{arXiv preprint arXiv:2107.03374}, 2021.

\bibitem[Chen et~al.(2022)Chen, Ge, Tong, Wang, Song, Wang, and Luo]{DBLP:conf/nips/ChenGTWSWL22}
Shoufa Chen, Chongjian Ge, Zhan Tong, Jiangliu Wang, Yibing Song, Jue Wang, and Ping Luo.
\newblock Adaptformer: Adapting vision transformers for scalable visual recognition.
\newblock In \emph{Advances in Neural Information Processing Systems 35}, 2022.

\bibitem[Cheng et~al.(2017)Cheng, Han, and Lu]{cheng2017remote}
Gong Cheng, Junwei Han, and Xiaoqiang Lu.
\newblock Remote sensing image scene classification: Benchmark and state of the art.
\newblock \emph{Proceedings of the IEEE}, 105\penalty0 (10):\penalty0 1865--1883, 2017.

\bibitem[Cimpoi et~al.(2014)Cimpoi, Maji, Kokkinos, Mohamed, and Vedaldi]{cimpoi2014describing}
Mircea Cimpoi, Subhransu Maji, Iasonas Kokkinos, Sammy Mohamed, and Andrea Vedaldi.
\newblock Describing textures in the wild.
\newblock In \emph{Proceedings of the IEEE conference on computer vision and pattern recognition}, pp.\  3606--3613, 2014.

\bibitem[Clark et~al.(2018)Clark, Cowhey, Etzioni, Khot, Sabharwal, Schoenick, and Tafjord]{clark2018think}
Peter Clark, Isaac Cowhey, Oren Etzioni, Tushar Khot, Ashish Sabharwal, Carissa Schoenick, and Oyvind Tafjord.
\newblock Think you have solved question answering? try arc, the ai2 reasoning challenge.
\newblock \emph{arXiv preprint arXiv:1803.05457}, 2018.

\bibitem[Cobbe et~al.(2021)Cobbe, Kosaraju, Bavarian, Chen, Jun, Kaiser, Plappert, Tworek, Hilton, Nakano, et~al.]{cobbe2021training}
Karl Cobbe, Vineet Kosaraju, Mohammad Bavarian, Mark Chen, Heewoo Jun, Lukasz Kaiser, Matthias Plappert, Jerry Tworek, Jacob Hilton, Reiichiro Nakano, et~al.
\newblock Training verifiers to solve math word problems.
\newblock \emph{arXiv preprint arXiv:2110.14168}, 2021.

\bibitem[Cui et~al.(2024)Cui, Yuan, Ding, Yao, He, Zhu, Ni, Xie, Xie, Lin, et~al.]{cui2024ultrafeedback}
Ganqu Cui, Lifan Yuan, Ning Ding, Guanming Yao, Bingxiang He, Wei Zhu, Yuan Ni, Guotong Xie, Ruobing Xie, Yankai Lin, et~al.
\newblock Ultrafeedback: Boosting language models with scaled ai feedback.
\newblock In \emph{Forty-first International Conference on Machine Learning}, 2024.

\bibitem[Deep et~al.(2024)Deep, Bhardwaj, and Poria]{DBLP:journals/corr/abs-2406-11617}
Pala~Tej Deep, Rishabh Bhardwaj, and Soujanya Poria.
\newblock Della-merging: Reducing interference in model merging through magnitude-based sampling.
\newblock \emph{CoRR}, abs/2406.11617, 2024.

\bibitem[Devlin et~al.(2019)Devlin, Chang, Lee, and Toutanova]{DBLP:conf/naacl/DevlinCLT19}
Jacob Devlin, Ming{-}Wei Chang, Kenton Lee, and Kristina Toutanova.
\newblock {BERT:} pre-training of deep bidirectional transformers for language understanding.
\newblock In \emph{Proceedings of the 2019 Conference of the North American Chapter of the Association for Computational Linguistics: Human Language Technologies}, pp.\  4171--4186. Association for Computational Linguistics, 2019.

\bibitem[Dodge et~al.(2020)Dodge, Ilharco, Schwartz, Farhadi, Hajishirzi, and Smith]{DBLP:journals/corr/abs-2002-06305}
Jesse Dodge, Gabriel Ilharco, Roy Schwartz, Ali Farhadi, Hannaneh Hajishirzi, and Noah~A. Smith.
\newblock Fine-tuning pretrained language models: Weight initializations, data orders, and early stopping.
\newblock \emph{CoRR}, abs/2002.06305, 2020.

\bibitem[Dosovitskiy et~al.(2021)Dosovitskiy, Beyer, Kolesnikov, Weissenborn, Zhai, Unterthiner, Dehghani, Minderer, Heigold, Gelly, Uszkoreit, and Houlsby]{DBLP:conf/iclr/DosovitskiyB0WZ21}
Alexey Dosovitskiy, Lucas Beyer, Alexander Kolesnikov, Dirk Weissenborn, Xiaohua Zhai, Thomas Unterthiner, Mostafa Dehghani, Matthias Minderer, Georg Heigold, Sylvain Gelly, Jakob Uszkoreit, and Neil Houlsby.
\newblock An image is worth 16x16 words: Transformers for image recognition at scale.
\newblock In \emph{9th International Conference on Learning Representations}. OpenReview.net, 2021.

\bibitem[Dubey et~al.(2024)Dubey, Jauhri, Pandey, Kadian, Al{-}Dahle, Letman, Mathur, Schelten, Yang, Fan, Goyal, Hartshorn, Yang, Mitra, Sravankumar, Korenev, Hinsvark, Rao, Zhang, Rodriguez, Gregerson, Spataru, Rozi{\`{e}}re, Biron, Tang, Chern, Caucheteux, Nayak, Bi, Marra, McConnell, Keller, Touret, Wu, Wong, Ferrer, Nikolaidis, Allonsius, Song, Pintz, Livshits, Esiobu, Choudhary, Mahajan, Garcia{-}Olano, Perino, Hupkes, Lakomkin, AlBadawy, Lobanova, Dinan, Smith, Radenovic, Zhang, Synnaeve, Lee, Anderson, Nail, Mialon, Pang, Cucurell, Nguyen, Korevaar, Xu, Touvron, Zarov, Ibarra, Kloumann, Misra, Evtimov, Copet, Lee, Geffert, Vranes, Park, Mahadeokar, Shah, van~der Linde, Billock, Hong, Lee, Fu, Chi, Huang, Liu, Wang, Yu, Bitton, Spisak, Park, Rocca, Johnstun, Saxe, Jia, Alwala, Upasani, Plawiak, Li, Heafield, Stone, and et~al.]{DBLP:journals/corr/abs-2407-21783}
Abhimanyu Dubey, Abhinav Jauhri, Abhinav Pandey, Abhishek Kadian, Ahmad Al{-}Dahle, Aiesha Letman, Akhil Mathur, Alan Schelten, Amy Yang, Angela Fan, Anirudh Goyal, Anthony Hartshorn, Aobo Yang, Archi Mitra, Archie Sravankumar, Artem Korenev, Arthur Hinsvark, Arun Rao, Aston Zhang, Aur{\'{e}}lien Rodriguez, Austen Gregerson, Ava Spataru, Baptiste Rozi{\`{e}}re, Bethany Biron, Binh Tang, Bobbie Chern, Charlotte Caucheteux, Chaya Nayak, Chloe Bi, Chris Marra, Chris McConnell, Christian Keller, Christophe Touret, Chunyang Wu, Corinne Wong, Cristian~Canton Ferrer, Cyrus Nikolaidis, Damien Allonsius, Daniel Song, Danielle Pintz, Danny Livshits, David Esiobu, Dhruv Choudhary, Dhruv Mahajan, Diego Garcia{-}Olano, Diego Perino, Dieuwke Hupkes, Egor Lakomkin, Ehab AlBadawy, Elina Lobanova, Emily Dinan, Eric~Michael Smith, Filip Radenovic, Frank Zhang, Gabriel Synnaeve, Gabrielle Lee, Georgia~Lewis Anderson, Graeme Nail, Gr{\'{e}}goire Mialon, Guan Pang, Guillem Cucurell, Hailey Nguyen, Hannah Korevaar, Hu~Xu, Hugo
  Touvron, Iliyan Zarov, Imanol~Arrieta Ibarra, Isabel~M. Kloumann, Ishan Misra, Ivan Evtimov, Jade Copet, Jaewon Lee, Jan Geffert, Jana Vranes, Jason Park, Jay Mahadeokar, Jeet Shah, Jelmer van~der Linde, Jennifer Billock, Jenny Hong, Jenya Lee, Jeremy Fu, Jianfeng Chi, Jianyu Huang, Jiawen Liu, Jie Wang, Jiecao Yu, Joanna Bitton, Joe Spisak, Jongsoo Park, Joseph Rocca, Joshua Johnstun, Joshua Saxe, Junteng Jia, Kalyan~Vasuden Alwala, Kartikeya Upasani, Kate Plawiak, Ke~Li, Kenneth Heafield, Kevin Stone, and et~al.
\newblock The llama 3 herd of models.
\newblock \emph{CoRR}, abs/2407.21783, 2024.

\bibitem[Ethayarajh et~al.(2024)Ethayarajh, Xu, Muennighoff, Jurafsky, and Kiela]{DBLP:journals/corr/abs-2402-01306}
Kawin Ethayarajh, Winnie Xu, Niklas Muennighoff, Dan Jurafsky, and Douwe Kiela.
\newblock {KTO:} model alignment as prospect theoretic optimization.
\newblock In \emph{International Conference on Machine Learning}. PMLR, 2024.

\bibitem[Han et~al.(2024)Han, Gao, Liu, Zhang, and Zhang]{DBLP:journals/corr/abs-2403-14608}
Zeyu Han, Chao Gao, Jinyang Liu, Jeff Zhang, and Sai~Qian Zhang.
\newblock Parameter-efficient fine-tuning for large models: {A} comprehensive survey.
\newblock \emph{CoRR}, abs/2403.14608, 2024.

\bibitem[He et~al.(2023)He, Li, Zhang, Yang, and Wang]{DBLP:conf/aaai/HeLZYW23}
Xuehai He, Chunyuan Li, Pengchuan Zhang, Jianwei Yang, and Xin~Eric Wang.
\newblock Parameter-efficient model adaptation for vision transformers.
\newblock In \emph{Thirty-Seventh {AAAI} Conference on Artificial Intelligence}, pp.\  817--825. {AAAI} Press, 2023.

\bibitem[Helber et~al.(2019)Helber, Bischke, Dengel, and Borth]{helber2019eurosat}
Patrick Helber, Benjamin Bischke, Andreas Dengel, and Damian Borth.
\newblock Eurosat: A novel dataset and deep learning benchmark for land use and land cover classification.
\newblock \emph{IEEE Journal of Selected Topics in Applied Earth Observations and Remote Sensing}, 12\penalty0 (7):\penalty0 2217--2226, 2019.

\bibitem[Hendrycks et~al.(2020)Hendrycks, Burns, Basart, Zou, Mazeika, Song, and Steinhardt]{hendrycks2020measuring}
Dan Hendrycks, Collin Burns, Steven Basart, Andy Zou, Mantas Mazeika, Dawn Song, and Jacob Steinhardt.
\newblock Measuring massive multitask language understanding.
\newblock \emph{arXiv preprint arXiv:2009.03300}, 2020.

\bibitem[Houlsby et~al.(2019)Houlsby, Giurgiu, Jastrzebski, Morrone, de~Laroussilhe, Gesmundo, Attariyan, and Gelly]{DBLP:conf/icml/HoulsbyGJMLGAG19}
Neil Houlsby, Andrei Giurgiu, Stanislaw Jastrzebski, Bruna Morrone, Quentin de~Laroussilhe, Andrea Gesmundo, Mona Attariyan, and Sylvain Gelly.
\newblock Parameter-efficient transfer learning for {NLP}.
\newblock In \emph{Proceedings of the 36th International Conference on Machine Learning}, volume~97 of \emph{Proceedings of Machine Learning Research}, pp.\  2790--2799. {PMLR}, 2019.

\bibitem[Hu et~al.(2022)Hu, Shen, Wallis, Allen{-}Zhu, Li, Wang, Wang, and Chen]{DBLP:conf/iclr/HuSWALWWC22}
Edward~J. Hu, Yelong Shen, Phillip Wallis, Zeyuan Allen{-}Zhu, Yuanzhi Li, Shean Wang, Lu~Wang, and Weizhu Chen.
\newblock Lora: Low-rank adaptation of large language models.
\newblock In \emph{The Tenth International Conference on Learning Representations}. OpenReview.net, 2022.

\bibitem[Ilharco et~al.(2023)Ilharco, Ribeiro, Wortsman, Schmidt, Hajishirzi, and Farhadi]{DBLP:conf/iclr/IlharcoRWSHF23}
Gabriel Ilharco, Marco~T{\'{u}}lio Ribeiro, Mitchell Wortsman, Ludwig Schmidt, Hannaneh Hajishirzi, and Ali Farhadi.
\newblock Editing models with task arithmetic.
\newblock In \emph{The Eleventh International Conference on Learning Representations}. OpenReview.net, 2023.

\bibitem[Jiang et~al.(2023)Jiang, Sablayrolles, Mensch, Bamford, Chaplot, de~Las~Casas, Bressand, Lengyel, Lample, Saulnier, Lavaud, Lachaux, Stock, Scao, Lavril, Wang, Lacroix, and Sayed]{DBLP:journals/corr/abs-2310-06825}
Albert~Q. Jiang, Alexandre Sablayrolles, Arthur Mensch, Chris Bamford, Devendra~Singh Chaplot, Diego de~Las~Casas, Florian Bressand, Gianna Lengyel, Guillaume Lample, Lucile Saulnier, L{\'{e}}lio~Renard Lavaud, Marie{-}Anne Lachaux, Pierre Stock, Teven~Le Scao, Thibaut Lavril, Thomas Wang, Timoth{\'{e}}e Lacroix, and William~El Sayed.
\newblock Mistral 7b.
\newblock \emph{CoRR}, abs/2310.06825, 2023.

\bibitem[Klema \& Laub(1980)Klema and Laub]{klema1980singular}
Virginia Klema and Alan Laub.
\newblock The singular value decomposition: Its computation and some applications.
\newblock \emph{IEEE Transactions on automatic control}, 25\penalty0 (2):\penalty0 164--176, 1980.

\bibitem[Krause et~al.(2013)Krause, Stark, Deng, and Fei-Fei]{krause20133d}
Jonathan Krause, Michael Stark, Jia Deng, and Li~Fei-Fei.
\newblock 3d object representations for fine-grained categorization.
\newblock In \emph{Proceedings of the IEEE international conference on computer vision workshops}, pp.\  554--561, 2013.

\bibitem[LeCun et~al.(2010)LeCun, Cortes, and Burges]{lecun2010mnist}
Yann LeCun, Corinna Cortes, and CJ~Burges.
\newblock Mnist handwritten digit database.
\newblock \emph{ATT Labs [Online]. Available: http://yann.lecun.com/exdb/mnist}, 2, 2010.

\bibitem[Li \& Liang(2021)Li and Liang]{DBLP:conf/acl/LiL20}
Xiang~Lisa Li and Percy Liang.
\newblock Prefix-tuning: Optimizing continuous prompts for generation.
\newblock In \emph{Proceedings of the 59th Annual Meeting of the Association for Computational Linguistics and the 11th International Joint Conference on Natural Language Processing}, pp.\  4582--4597. Association for Computational Linguistics, 2021.

\bibitem[Lin et~al.(2021)Lin, Hilton, and Evans]{lin2021truthfulqa}
Stephanie Lin, Jacob Hilton, and Owain Evans.
\newblock Truthfulqa: Measuring how models mimic human falsehoods.
\newblock \emph{arXiv preprint arXiv:2109.07958}, 2021.

\bibitem[Liu et~al.(2024)Liu, Xiao, Li, Lee, Han, Dao, and Cai]{DBLP:journals/corr/abs-2402-10193}
James Liu, Guangxuan Xiao, Kai Li, Jason~D. Lee, Song Han, Tri Dao, and Tianle Cai.
\newblock Bitdelta: Your fine-tune may only be worth one bit.
\newblock \emph{CoRR}, abs/2402.10193, 2024.

\bibitem[Liu et~al.(2021)Liu, Lin, Cao, Hu, Wei, Zhang, Lin, and Guo]{DBLP:conf/iccv/LiuL00W0LG21}
Ze~Liu, Yutong Lin, Yue Cao, Han Hu, Yixuan Wei, Zheng Zhang, Stephen Lin, and Baining Guo.
\newblock Swin transformer: Hierarchical vision transformer using shifted windows.
\newblock In \emph{2021 {IEEE/CVF} International Conference on Computer Vision}, pp.\  9992--10002. {IEEE}, 2021.

\bibitem[Lu et~al.(2024)Lu, Fan, Wei, Qu, Chen, and Cheng]{DBLP:journals/corr/abs-2406-15479}
Zhenyi Lu, Chenghao Fan, Wei Wei, Xiaoye Qu, Dangyang Chen, and Yu~Cheng.
\newblock Twin-merging: Dynamic integration of modular expertise in model merging.
\newblock \emph{CoRR}, abs/2406.15479, 2024.

\bibitem[Luo et~al.(2023)Luo, Sun, Xu, Zhao, Lou, Tao, Geng, Lin, Chen, and Zhang]{DBLP:journals/corr/abs-2308-09583}
Haipeng Luo, Qingfeng Sun, Can Xu, Pu~Zhao, Jianguang Lou, Chongyang Tao, Xiubo Geng, Qingwei Lin, Shifeng Chen, and Dongmei Zhang.
\newblock Wizardmath: Empowering mathematical reasoning for large language models via reinforced evol-instruct.
\newblock \emph{CoRR}, abs/2308.09583, 2023.

\bibitem[Netzer et~al.(2011)Netzer, Wang, Coates, Bissacco, Wu, Ng, et~al.]{netzer2011reading}
Yuval Netzer, Tao Wang, Adam Coates, Alessandro Bissacco, Baolin Wu, Andrew~Y Ng, et~al.
\newblock Reading digits in natural images with unsupervised feature learning.
\newblock In \emph{NIPS workshop on deep learning and unsupervised feature learning}, volume 2011, pp.\ ~4. Granada, 2011.

\bibitem[Radford et~al.(2018)Radford, Narasimhan, Salimans, Sutskever, et~al.]{radford2018improving}
Alec Radford, Karthik Narasimhan, Tim Salimans, Ilya Sutskever, et~al.
\newblock Improving language understanding by generative pre-training.
\newblock 2018.

\bibitem[Radford et~al.(2021)Radford, Kim, Hallacy, Ramesh, Goh, Agarwal, Sastry, Askell, Mishkin, Clark, et~al.]{radford2021learning}
Alec Radford, Jong~Wook Kim, Chris Hallacy, Aditya Ramesh, Gabriel Goh, Sandhini Agarwal, Girish Sastry, Amanda Askell, Pamela Mishkin, Jack Clark, et~al.
\newblock Learning transferable visual models from natural language supervision.
\newblock In \emph{International conference on machine learning}, pp.\  8748--8763. PMLR, 2021.

\bibitem[Rafailov et~al.(2023)Rafailov, Sharma, Mitchell, Manning, Ermon, and Finn]{DBLP:conf/nips/RafailovSMMEF23}
Rafael Rafailov, Archit Sharma, Eric Mitchell, Christopher~D. Manning, Stefano Ermon, and Chelsea Finn.
\newblock Direct preference optimization: Your language model is secretly a reward model.
\newblock In \emph{Advances in Neural Information Processing Systems 36}, 2023.

\bibitem[Sakaguchi et~al.(2021)Sakaguchi, Bras, Bhagavatula, and Choi]{sakaguchi2021winogrande}
Keisuke Sakaguchi, Ronan~Le Bras, Chandra Bhagavatula, and Yejin Choi.
\newblock Winogrande: An adversarial winograd schema challenge at scale.
\newblock \emph{Communications of the ACM}, 64\penalty0 (9):\penalty0 99--106, 2021.

\bibitem[Sandler et~al.(2022)Sandler, Zhmoginov, Vladymyrov, and Jackson]{DBLP:conf/cvpr/0002ZV022}
Mark Sandler, Andrey Zhmoginov, Max Vladymyrov, and Andrew Jackson.
\newblock Fine-tuning image transformers using learnable memory.
\newblock In \emph{{IEEE/CVF} Conference on Computer Vision and Pattern Recognition, {CVPR} 2022, New Orleans, LA, USA, June 18-24, 2022}, pp.\  12145--12154. {IEEE}, 2022.

\bibitem[Stallkamp et~al.(2011)Stallkamp, Schlipsing, Salmen, and Igel]{stallkamp2011german}
Johannes Stallkamp, Marc Schlipsing, Jan Salmen, and Christian Igel.
\newblock The german traffic sign recognition benchmark: a multi-class classification competition.
\newblock In \emph{The 2011 international joint conference on neural networks}, pp.\  1453--1460. IEEE, 2011.

\bibitem[Tong et~al.(2024)Tong, Zhang, Wang, Wu, and He]{DBLP:journals/corr/abs-2407-13690}
Yuxuan Tong, Xiwen Zhang, Rui Wang, Ruidong Wu, and Junxian He.
\newblock Dart-math: Difficulty-aware rejection tuning for mathematical problem-solving.
\newblock \emph{CoRR}, abs/2407.13690, 2024.

\bibitem[Xiao et~al.(2016)Xiao, Ehinger, Hays, Torralba, and Oliva]{xiao2016sun}
Jianxiong Xiao, Krista~A Ehinger, James Hays, Antonio Torralba, and Aude Oliva.
\newblock Sun database: Exploring a large collection of scene categories.
\newblock \emph{International Journal of Computer Vision}, 119:\penalty0 3--22, 2016.

\bibitem[Xin et~al.(2024)Xin, Luo, Zhou, Du, Liu, Fan, Li, and Du]{DBLP:journals/corr/abs-2402-02242}
Yi~Xin, Siqi Luo, Haodi Zhou, Junlong Du, Xiaohong Liu, Yue Fan, Qing Li, and Yuntao Du.
\newblock Parameter-efficient fine-tuning for pre-trained vision models: {A} survey.
\newblock \emph{CoRR}, abs/2402.02242, 2024.

\bibitem[Yadav et~al.(2023)Yadav, Tam, Choshen, Raffel, and Bansal]{DBLP:conf/nips/YadavTCRB23}
Prateek Yadav, Derek Tam, Leshem Choshen, Colin~A. Raffel, and Mohit Bansal.
\newblock Ties-merging: Resolving interference when merging models.
\newblock In \emph{Advances in Neural Information Processing Systems 36}, 2023.

\bibitem[Yu et~al.(2024)Yu, Yu, Yu, Huang, and Li]{yu2023language}
Le~Yu, Bowen Yu, Haiyang Yu, Fei Huang, and Yongbin Li.
\newblock Language models are super mario: Absorbing abilities from homologous models as a free lunch.
\newblock In \emph{International Conference on Machine Learning}. PMLR, 2024.

\bibitem[Zellers et~al.(2019)Zellers, Holtzman, Bisk, Farhadi, and Choi]{zellers2019hellaswag}
Rowan Zellers, Ari Holtzman, Yonatan Bisk, Ali Farhadi, and Yejin Choi.
\newblock Hellaswag: Can a machine really finish your sentence?
\newblock \emph{arXiv preprint arXiv:1905.07830}, 2019.

\bibitem[Zhao et~al.(2023)Zhao, Zhou, Li, Tang, Wang, Hou, Min, Zhang, Zhang, Dong, Du, Yang, Chen, Chen, Jiang, Ren, Li, Tang, Liu, Liu, Nie, and Wen]{DBLP:journals/corr/abs-2303-18223}
Wayne~Xin Zhao, Kun Zhou, Junyi Li, Tianyi Tang, Xiaolei Wang, Yupeng Hou, Yingqian Min, Beichen Zhang, Junjie Zhang, Zican Dong, Yifan Du, Chen Yang, Yushuo Chen, Zhipeng Chen, Jinhao Jiang, Ruiyang Ren, Yifan Li, Xinyu Tang, Zikang Liu, Peiyu Liu, Jian{-}Yun Nie, and Ji{-}Rong Wen.
\newblock A survey of large language models.
\newblock \emph{CoRR}, abs/2303.18223, 2023.

\bibitem[Zheng et~al.(2024)Zheng, Wang, Ji, Huang, and Peng]{DBLP:journals/corr/abs-2404-16792}
Chujie Zheng, Ziqi Wang, Heng Ji, Minlie Huang, and Nanyun Peng.
\newblock Weak-to-strong extrapolation expedites alignment.
\newblock \emph{CoRR}, abs/2404.16792, 2024.

\bibitem[Zhou et~al.(2023)Zhou, Lu, Mishra, Brahma, Basu, Luan, Zhou, and Hou]{zhou2023instructionfollowingevaluationlargelanguage}
Jeffrey Zhou, Tianjian Lu, Swaroop Mishra, Siddhartha Brahma, Sujoy Basu, Yi~Luan, Denny Zhou, and Le~Hou.
\newblock Instruction-following evaluation for large language models, 2023.
\newblock URL \url{https://arxiv.org/abs/2311.07911}.

\end{thebibliography}
\bibliographystyle{iclr2025_conference}
\newpage
\appendix
\label{section-appendix}

\section{Experimental Details}
\label{appendix:settings}

For the loss estimation experiments, we set the constant $C = 5$ to calculate the approximation term. We interpolate the $\Delta \widetilde{\bm{W}}$ parameter at values of 0.2, 0.4, 0.6, and 0.8 to generate different models. For all large language model evaluations in Chapters 4 and 5, we employ the lm-eval framework for assessment. Further details on the tested models and datasets can be found in the original paper.

\section{Full Experimental Results}
\subsection{Extension of DARE}
\label{appendix:dare}
\begin{figure}[htbp]
  \centering
  \includegraphics[width=1.0\columnwidth]{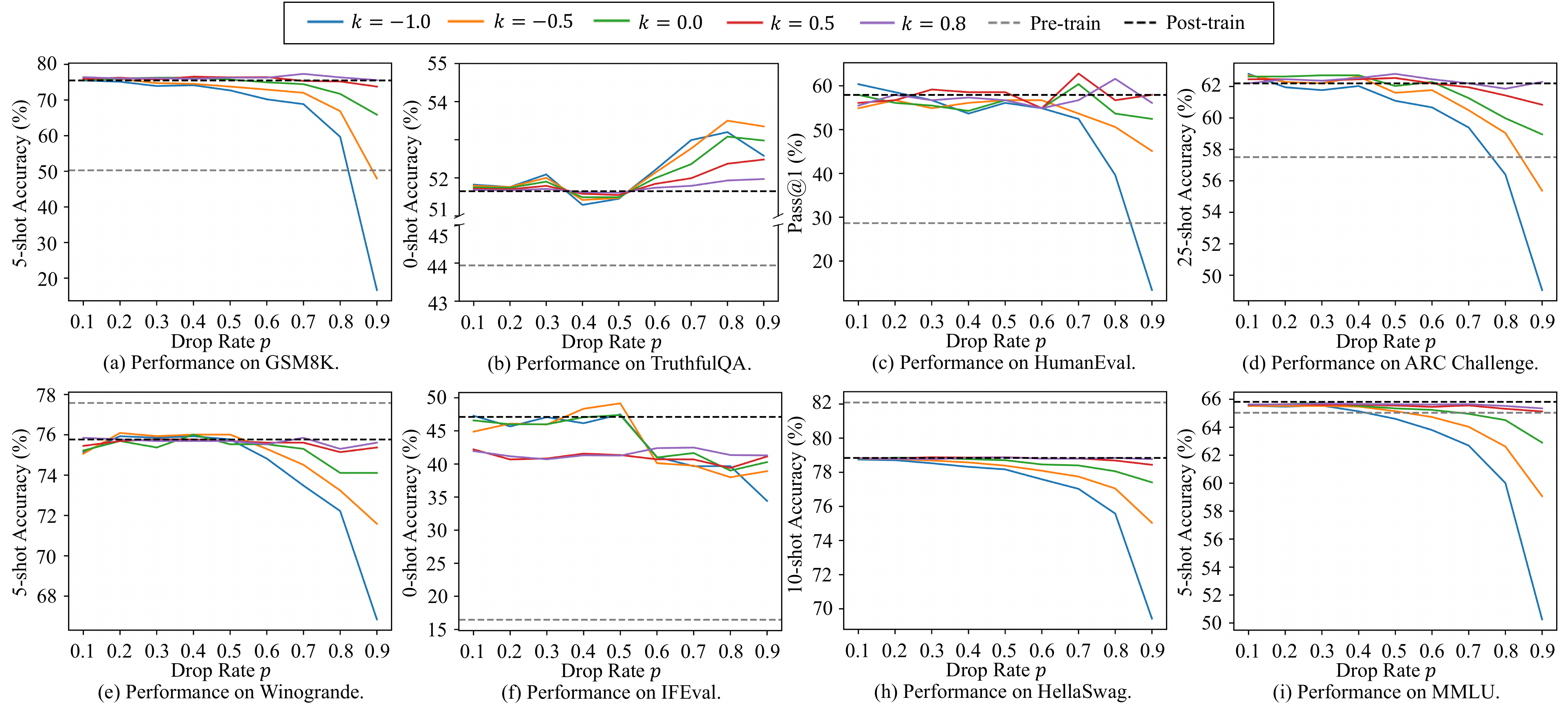} 
  \caption{The performance of LLaMA3-8B-Instruct on the all benchmarks under varying $p$ and $k$.}
  \vspace{-10px}\label{fig:app_dare_llama_all}
\end{figure}

\begin{figure}[htbp]
  \centering
  \includegraphics[width=1.0\columnwidth]{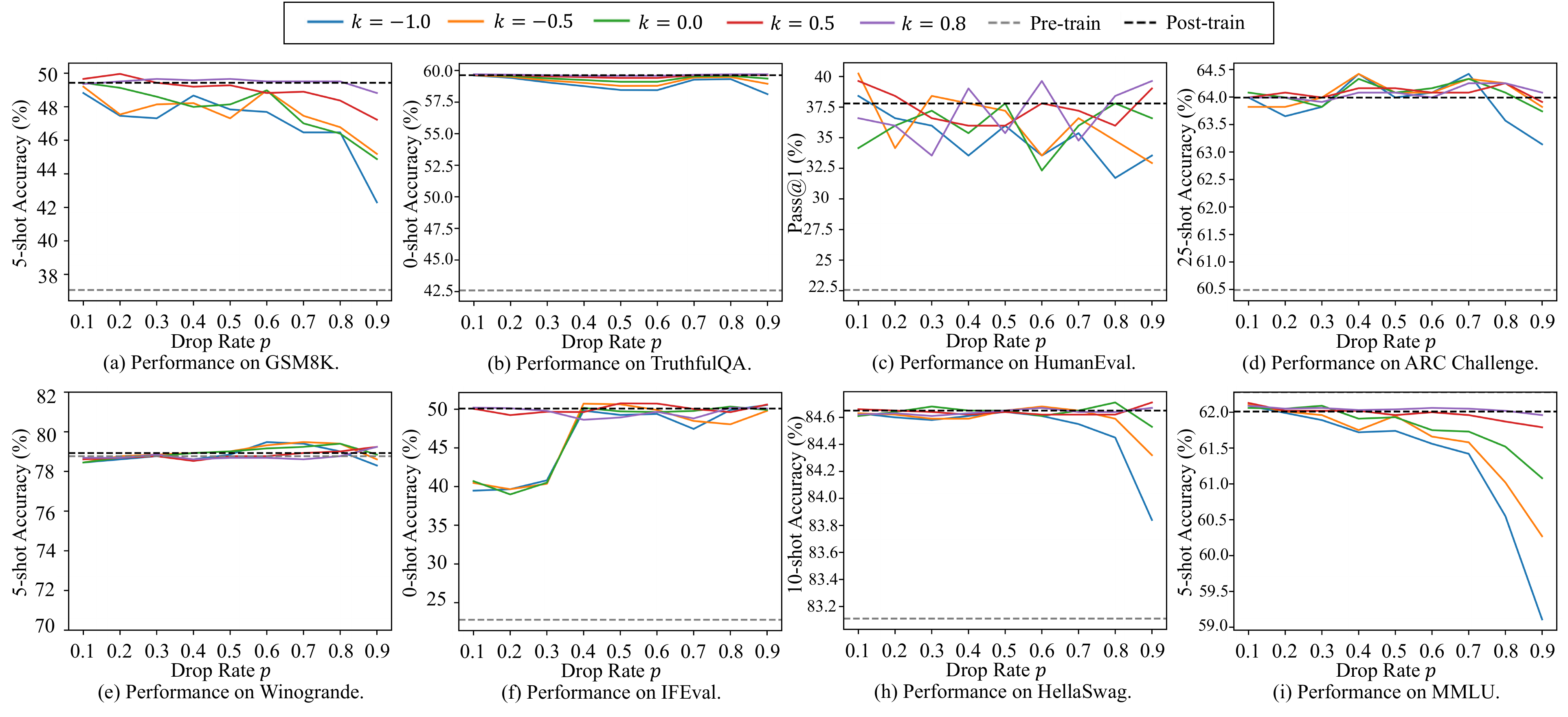} 
  \caption{The performance of Mistral-7B-Instruct-v0.3 on the all benchmarks under varying $p$ and $k$.}
  \vspace{-10px}\label{fig:app_dare_mistral_all}
\end{figure}

\begin{figure}[htbp]
  \centering
  \includegraphics[width=1.0\columnwidth]{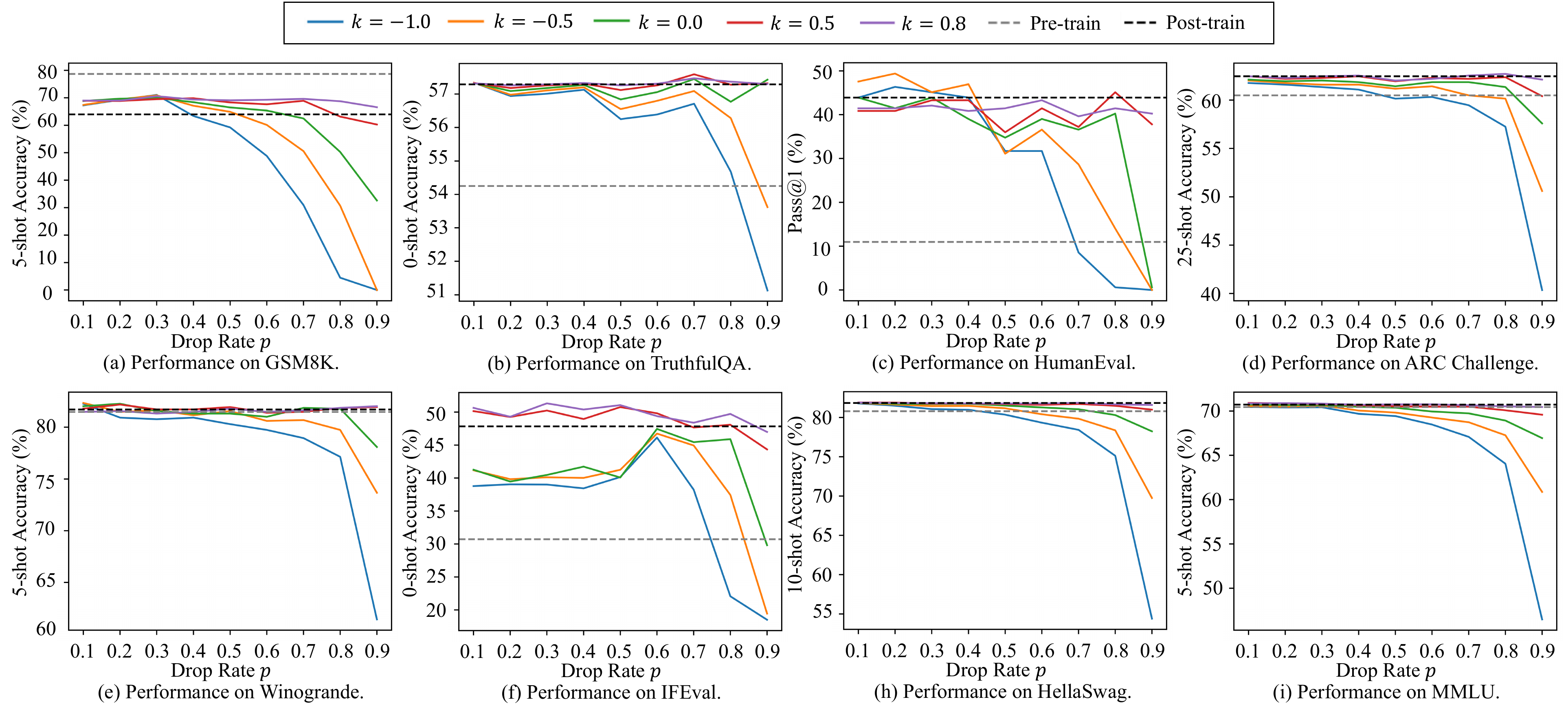} 
  \caption{The performance of Qwen2-7B-Instruct on the all benchmarks under varying $p$ and $k$.}
  \vspace{-10px}\label{fig:app_dare_qwen_all}
\end{figure}

\begin{figure}[htbp]
  \centering
  \includegraphics[width=1.0\columnwidth]{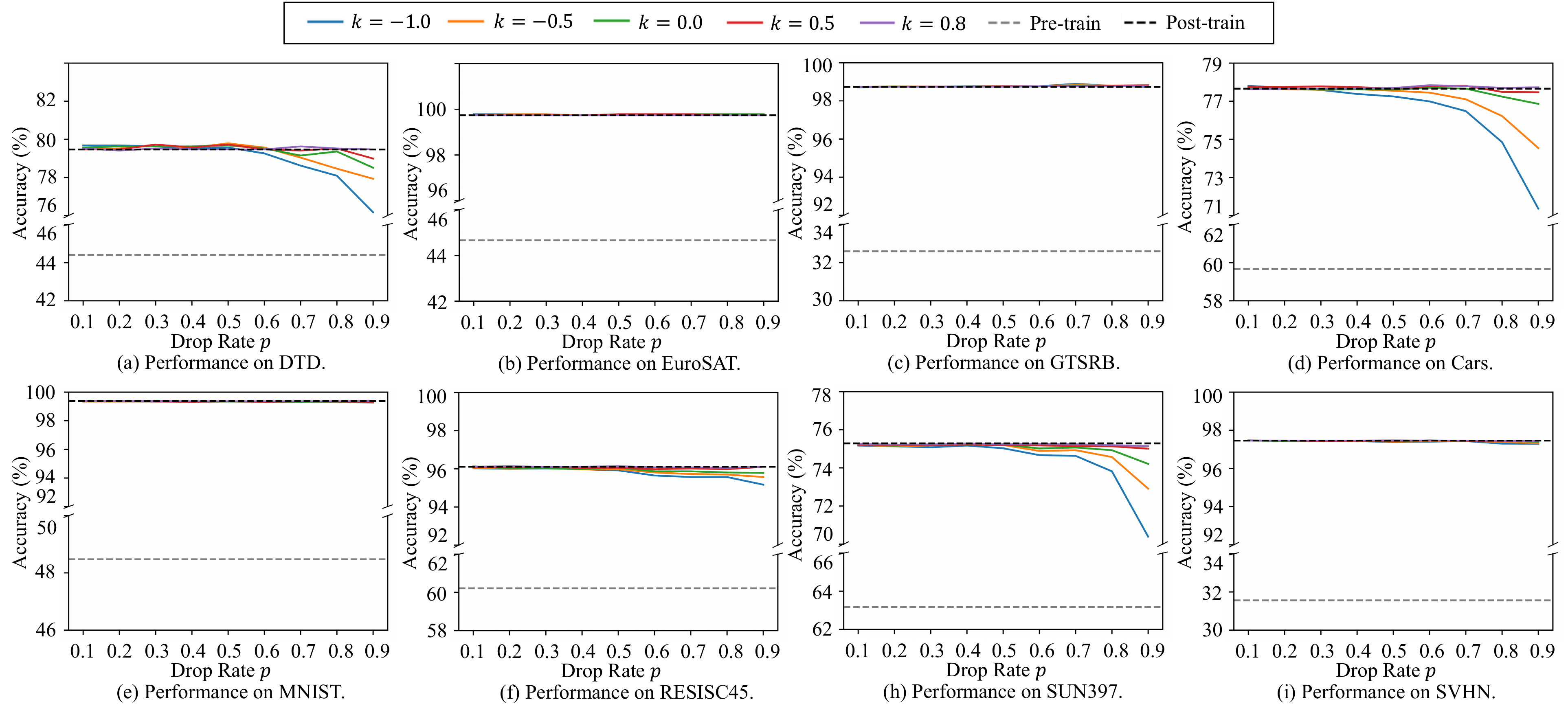} 
  \caption{The performance of ViT-B-32 on the all benchmarks under varying $p$ and $k$.}
  \vspace{-10px}\label{fig:app_dare_vit_all}
\end{figure}

We conduct a thorough experimental validation on the extension of DARE. The results of LLaMA3-8B-Instruct, Mistral-7B-Instruct-v0.3, and ViT-B-32 across eight benchmarks are presented in Figure~\ref{fig:app_dare_llama_all}, Figure~\ref{fig:app_dare_mistral_all}, Figure~\ref{fig:app_dare_qwen_all}, and Figure~\ref{fig:app_dare_vit_all}, respectively. 

\subsection{Extension of BitDelta}
\begin{figure}[htbp]
  \centering
  \includegraphics[width=1.0\columnwidth]{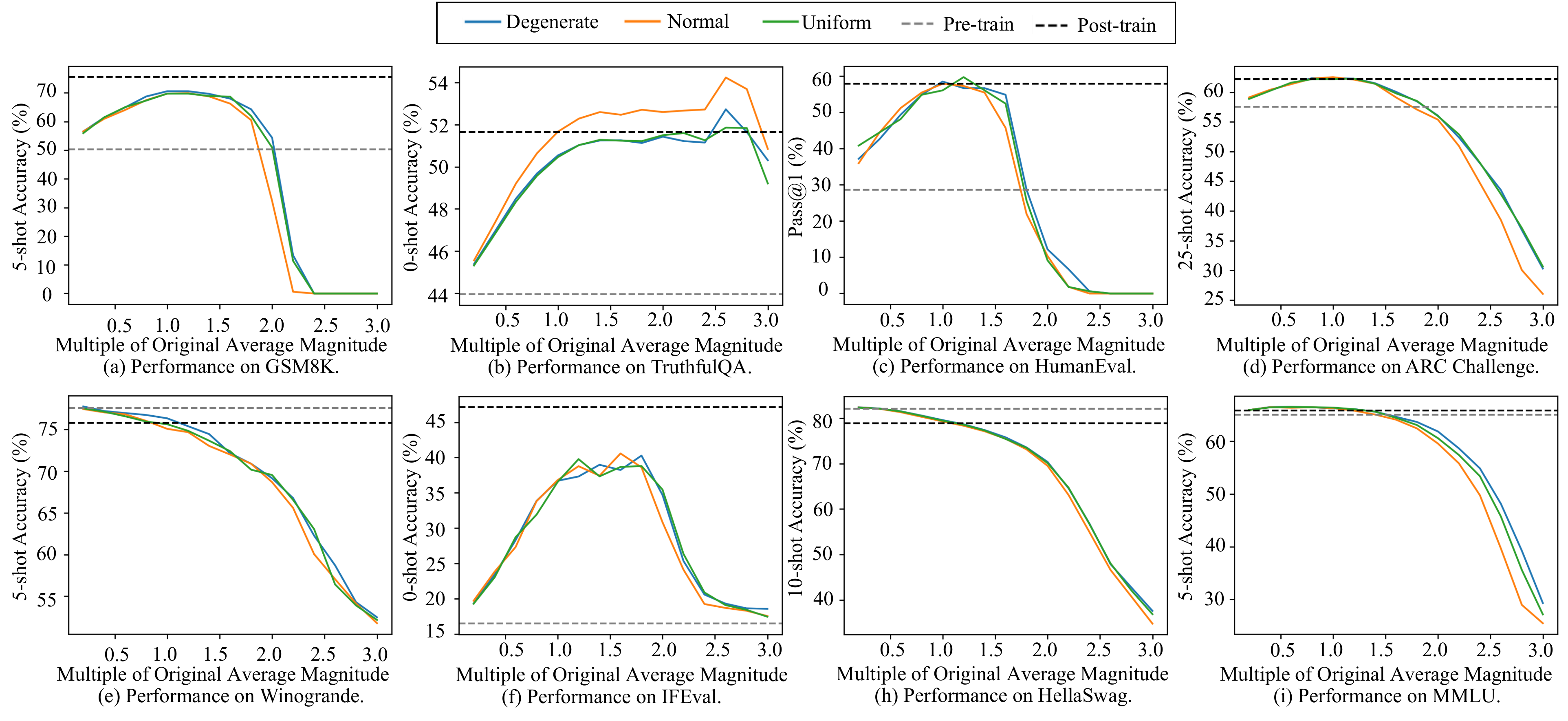} 
  \caption{Validation of the extension of BitDelta on LLaMA3-8B-Instruct.}
  \vspace{-10px}\label{fig:app_bitdelta_llama_all}
\end{figure}

The results of LLaMA3-8B-Instruct across eight benchmarks are presented in Figure~\ref{fig:app_bitdelta_llama_all}.

\subsection{Discussion on EXPO}
\label{appendix:expo_discuss}
\begin{figure}[htbp]
  \centering
  \includegraphics[width=1.0\columnwidth]{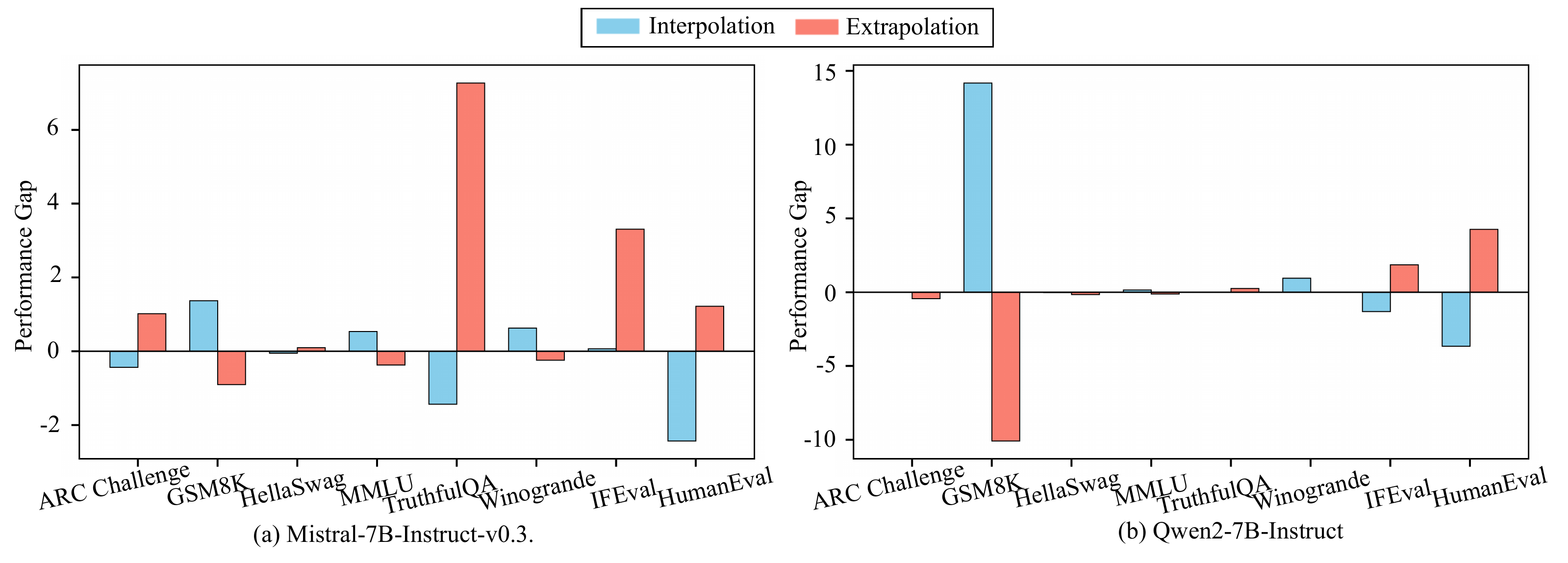} 
  \caption{Comparison of Extrapolation and Interpolation Performance.}
  \vspace{-10px}\label{fig:app_expo_discuss_all}
\end{figure}

In Figure~\ref{fig:app_expo_discuss_all}, we present the comparision of interpolation and extrapolation.

\end{document}